\documentclass[%
 aip,
 amsmath,amssymb,
 reprint,%
]{revtex4-1}

\usepackage{graphicx}
\usepackage{dcolumn}
\usepackage{bm}

\usepackage[colorlinks,allcolors=blue]{hyperref} 

\usepackage[utf8]{inputenc}
\usepackage[T1]{fontenc}
\usepackage{mathptmx}

\usepackage{subcaption}
\usepackage{caption}

\usepackage{calrsfs}
\DeclareMathAlphabet{\pazocal}{OMS}{zplm}{m}{n}
\SetMathAlphabet\pazocal{bold}{OMS}{zplm}{bx}{n}

\newcommand{\pzc}{\pazocal{A}}
\newcommand{\pzb}{\pazocal{B}}
\newcommand{\pzl}{\pazocal{L}}
\newcommand{\pzs}{\pazocal{S}}

\newcommand{\MATHUCC}{School of Mathematical Sciences, University College Cork, Cork T12 XF62, Ireland}
\newcommand{\RTRC}{Raytheon Technologies Research Center Ireland, Cork T23 XN53, Ireland}

\begin{document}

\preprint{AIP/123-QED}

\title[Multifunctionality in a Reservoir Computer]{Multifunctionality in a Reservoir Computer}

\author{Andrew Flynn}
\email{andrew\_flynn@umail.ucc.ie}
\affiliation{ 
\MATHUCC
}%

\author{Vassilios A. Tsachouridis}
\affiliation{%
\RTRC
}%

\author{Andreas Amann}
\affiliation{ 
\MATHUCC
}%

\date{\today}
\begin{abstract}
Multifunctionality is a well observed phenomenological feature of biological neural networks and considered to be of fundamental importance to the survival of certain species over time. These multifunctional neural networks are capable of performing more than one task without changing any network connections. In this paper we investigate how this neurological idiosyncrasy can be achieved in an artificial setting with a modern machine learning paradigm known as `Reservoir Computing'. A training technique is designed to enable a Reservoir Computer to perform tasks of a multifunctional nature. We explore the critical effects that changes in certain parameters can have on the Reservoir Computers' ability to express multifunctionality. We also expose the existence of several `untrained attractors'; attractors which dwell within the prediction state space of the Reservoir Computer that were not part of the training. We conduct a bifurcation analysis of these untrained attractors and discuss the implications of our results.
\end{abstract}

\maketitle

\begin{quotation}
Advancements in Machine Learning often arise from a `two-way street' between neuroscientific observation and mathematical representation. In this paper we stroll through such a street with inspiration from `Multifunctional Neural Networks'. These are networks of neurons whose activity patterns can change on the demand of performing a given duty but synapses remain fixed. We conceptualise multifunctionality in the context of Dynamical Systems and Machine Learning by using a Reservoir Computer as a means to realise this neurological feat in an artificial setting. More specifically, we train a Reservoir Computer to imitate the dynamics of numerous chaotic attractors from different sources, based on a given initial condition. To do this we design a training technique which `blends' and weights data from these chaotic attractors. We explore how different weightings and changes in the memory of this artificial neural network effect the desired learning outcomes. In doing so we uncover some `behind-the-scenes' bifurcations of several other attractors found to be lurking within the prediction state space that interfere with the networks capacity to express multifunctionality. Above all, this paper identifies some new application areas suitable to a Reservoir Computer and broadens the current understanding of the dynamical capabilities inherent to this learning system.
\end{quotation}

\section{\label{sec:Intro}Introduction}

Multifunctionality is an essential element of biological neural networks \cite{getting89Principles,Dickinson95MF,Marder96MFprinciples}. These multifunctional networks are distinct pools of neurons capable of performing a multitude of mutually exclusive tasks. To elaborate with example, it was found that a subset of the same bundle of neurons in the brain of the medicinal leech (\textit{Hirudo medicinalis}) can switch their activity pattern once it senses a change in its surroundings to drive either a swimming or crawling motion \cite{briggman2006imaging}. It is reported that a cluster of neurons located in the \textit{pre-B\"{o}tzinger complex} (a region of the mammalian brain stem), is responsible for regulating a switching between different respiratory patterns \cite{lieske00ReconfigurationBreathing}. Depending on a particular input, the neurons in this region of the brain can alter their activity pattern accordingly to elicit a switching between eupneic (regular) breathing, sighing or gasping. Furthermore, it is argued that multifunctionality in neural networks may naturally emerge from an efficient use of limited resources (in this case, neurons) and thus an evolutionary advantage in enduring environmental changes, reflecting the developmental history of certain organisms \cite{briggman08multifunctional}.

Nevertheless, what is ubiquitous amongst these multifunctional neural networks is that they in principle resemble a system with more than one modus operandi. Based on a particular input, there is a distinct activity pattern expressed by the neurons in the network in order to perform one of the many tasks required of it. When needed, these multifunctional neurons switch to another activity pattern to collectively execute a different task while the network connections remain fixed. Therefore, if an artificial neural network was trained to sustain more than one desired activity pattern it would in this sense be multifunctional. From a dynamical systems perspective, this type of behaviour is akin to a multistable system or a system with a coexistence of attractors. For further reading on multistable systems see \textcite{pisarchik14control_of_multistability}.

There is much to be gained in the pursuit of artificial intelligence by articulating our current knowledge of biological neural networks and dynamical systems in machine learning environments. In this paper we employ such a bilateral rationale to encapsulate multifunctionality in an artificial neural network by training a `Reservoir Computer'  \cite{Jaeger01ESN,Maass02_LSM,Verstraeten07RC} (RC) to facilitate the coexistence of more than one desired activity pattern.

The RC approach to machine learning has been successfully applied to a number of problems, for example, time series prediction \cite{JaegerHaas04ESN}, visual identification tasks \cite{jalalvand15_RC_VisualTasks}, real-time detection of epileptic seizures \cite{Buteneers13_Epilipsy}, and inferring from limited time series data; unmeasured state variables \cite{Lu17_UnmeasuredStatreVariables}, Lyapunov exponents \cite{Pathak17_Lyap_from_data}, and causal dependencies between variables \cite{Banerjee19CausalDep}. 

The basis of our research involves using a recent formulation of a continuous-time RC, presented by \textcite{LuHuntOtt18RC}. Here the RC was trained to perform short term predictions of a chaotic Lorenz system \cite{lorenz63model} and reconstruct the `climate' (qualitatively similar dynamical behaviour) of its famous butterfly shaped chaotic attractor. Taking this result into account, we train a RC to promote a coexistence of reconstructed chaotic attractors in its prediction state space, thus becoming multifunctional. In order to demonstrate the flexibility of our approach we consider training scenarios in which the climate of these chaotic attractors is reconstructed from a variety of sources. For example, we consider the case where these chaotic attractors are generated from two different systems entirely. We devise a training technique to `blend' data with a certain weighting parameter from these chaotic attractors. The choice of this weighting along with a parameter involved in tuning the memory of the RC is critical to achieving multifunctionality. We investigate the optimal setting of these parameters, from which we infer the regions in this parameter space where the RC achieves multifunctionality.


However, while we train the RC to realise more than one chaotic attractor in its prediction state space, we find several `untrained attractors' also residing here. These attractors inhabit the prediction state space but were not part of the training and limit the regions in which multifunctionality is obtained. A bifurcation analysis of these untrained attractors reveals some interesting dynamics where, for example, one of these attractors undergoes a period doubling route to chaos.

The structure of the rest of the paper is as follows. In Section \ref{sec:RC} we provide details of the RC approach to attractor reconstruction and present the training procedure we use to achieve multifunctionality in a RC. Next, in Section \ref{sec:MFinDS}, the problem of epitomising multifunctionality in a RC is further conceptualised. The trajectories on the chaotic attractors we use as our training data are also given here. In Section \ref{sec:Results} we present our main findings and then provide an extended discussion of our results in Section \ref{sec:DisConc}.

\section{\label{sec:RC}Reservoir Computing}

Echo-State Networks \cite{Jaeger01ESN} (ESNs) and Liquid-State Machines \cite{Maass02_LSM} (LSMs) are two independently proposed designs of artificial neural networks with recurrent connections that can be trained to provide a self-sustained activity pattern. While ESNs were engineered in the context of machine learning and LSMs were developed from a computational neuroscience perspective, they share a similar philosophy and as a result have become increasingly synonymous under the umbrella term of `Reservoir Computing' or indeed a `Reservoir Computer' (RC) \cite{Verstraeten07RC}. Both are based upon the notion that as long as the internal connections, or in this case the `reservoir', possess certain characteristics it is not necessary to adapt the internal weights of the network in order to achieve a desired learning outcome. Instead, it is sufficient to find an appropriate readout layer for a given task. Consequentially, as the internal connections of the reservoir remain unaltered throughout the training it can be represented physically. There are many interesting constructions of these `Physical Reservoir Computers', using, for example, an optoelectronic system \cite{Larger12_OptoElec}, or an octopus inspired soft robotic arm \cite{Nakajima15_InfoProc} . For a recent review see \textcite{tanaka19PhysRC}. 

For the purpose of keeping this paper self-contained, we now outline the anatomy and implementation stages of the RC setup we use that was proposed by \textcite{LuHuntOtt18RC}. We also present the technique we devised in order to train the RC to reconstruct the climate of more than one chaotic attractor based on a given initial condition. 


\subsection{\label{sec:ListenStage}Listening Stage}

\begin{figure}
    \centering
    \includegraphics[width=0.48\textwidth]{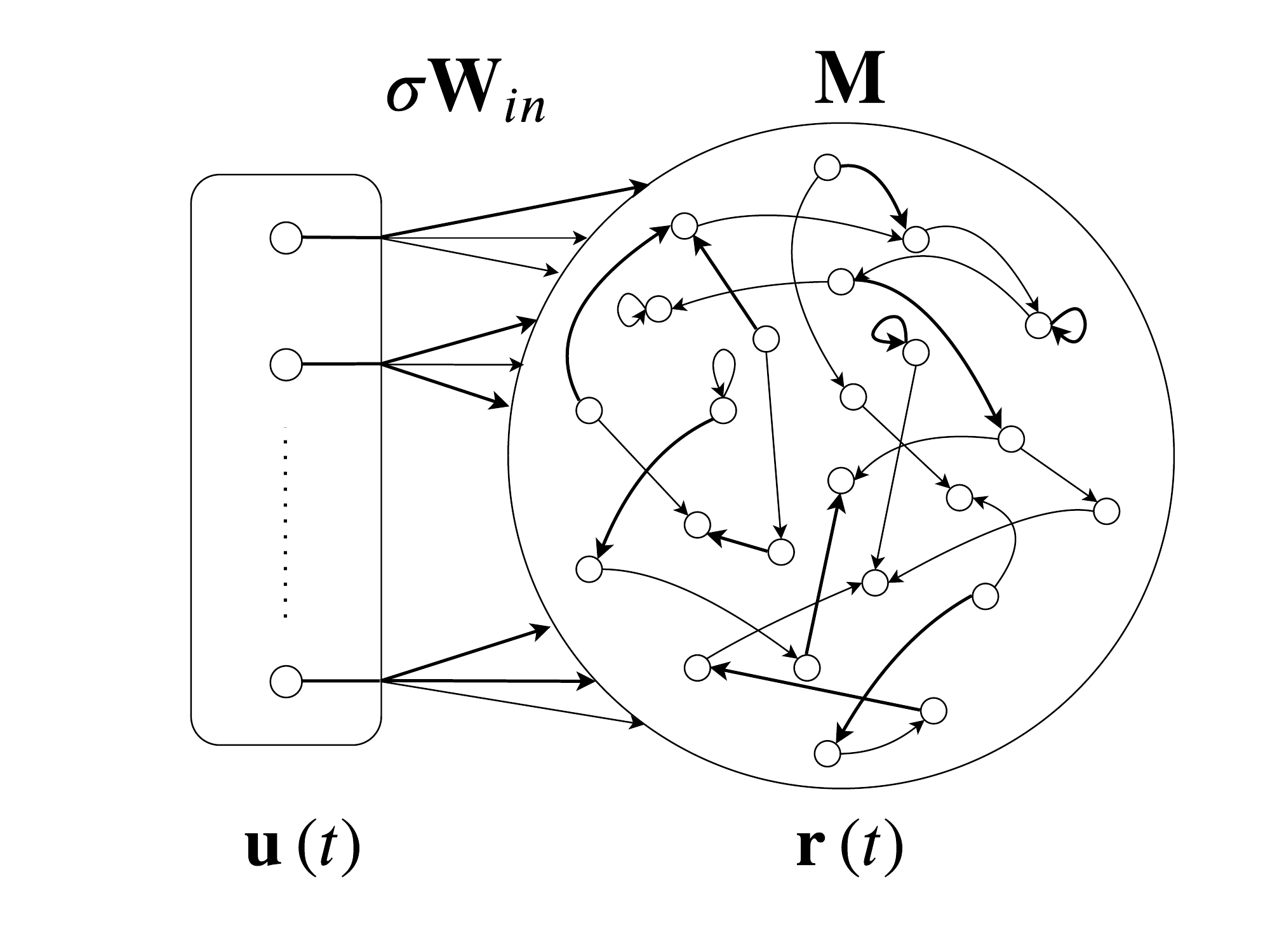}
    \caption{Listening reservoir: The input signal $\boldsymbol{u}(t)$ is projected by $\sigma \textbf{W}_{in}$ to drive a response from the reservoir state, $\boldsymbol{r}(t)$.}
    \label{fig:ListeningResFig}
\end{figure}

In the listening stage, as illustrated in Fig.\,\ref{fig:ListeningResFig}, the input data is used to drive the `listening reservoir' away from its initial state. This drive-response system evolves according to,
\begin{align}
    \dot{\boldsymbol{r}}(t) = \gamma \left[ - \boldsymbol{r}(t) + \tanh{\left( \,\, \textbf{M} \, \boldsymbol{r}(t) + \sigma \textbf{W}_{in} \, \boldsymbol{u}(t) \,\, \right)} \right]. \label{ListenRes}
\end{align}
Here $\boldsymbol{r}(t) \in \mathbb{R}^{N}$ is the state of the reservoir at a given time $t$ and $N$ denotes the number of artificial neurons. $\gamma$ is a parameter arising from the transformation of the discrete-time formulation in \textcite{JaegerHaas04ESN} to continuous-time. $\textbf{M} \in \mathbb{R}^{N \times N}$ is the adjacency matrix representing the internal network connections of the reservoir. The input strength parameter, $\sigma$, and $\textbf{W}_{in} \in \mathbb{R}^{N \times D}$, the input matrix, when multiplied together represent the weight given to the input vector $\boldsymbol{u}(t) \in \mathbb{R}^{D}$ as it is projected into the reservoir. Here $D$ is the dimension of the input data. We compute trajectories of Eq.\,\eqref{ListenRes} using the 4$^{th}$ order Runge-Kutta method with time step $\tau = 0.01$.

In the listening stage, we allow Eq.\,\eqref{ListenRes} to evolve from $t = 0$ to $t = t_{listen} = 200$ in order to remove any dependency the RC may have on its initial condition and synchronise to the input.

\subsection{\label{sec:Train}Training Stage}

In the training stage, we focus on the data generated from Eq.\,\eqref{ListenRes} and $\boldsymbol{u}(t)$ for $t_{listen} \leq t \leq t_{train} = 400$. The aim of training is to determine a post-processing function of the reservoir state, $\hat{\psi} \left( \boldsymbol{r} (t) \right)$, which can replace the input. Like in \textcite{LuHuntOtt18RC}, we consider post-processing functions of the following form,
\begin{align}
    \hat{\psi}(\boldsymbol{r}(t)) = \textbf{W}_{out} \, \boldsymbol{q}(\boldsymbol{r}(t)).
\end{align}
Here $\boldsymbol{q}(\boldsymbol{r}(t)) \in \mathbb{R}^{2 N}$ is a vector function which returns a vector where the first $N$ elements are $\boldsymbol{r}(t)$ and the second $N$ elements are $\boldsymbol{r}^{2}(t)$. While this step differs from many other manifestations of the readout layer, it is said to be advantageous over linear functions of $\boldsymbol{r}(t)$ and embeds further nonlinearity in the network. The `output matrix', $\textbf{W}_{out} \in \mathbb{R}^{D \times 2N}$ is determined by a ridge regression procedure which we now outline.

Each evaluation of $\boldsymbol{q}(\boldsymbol{r}(t))$ during the training time is stored in columns of the regularization matrix, $\textbf{X}$,
\begin{align}
    \textbf{X} = \left[ \begin{array}{cccc}
    \boldsymbol{q}(\boldsymbol{r}(t_{listen})) & \boldsymbol{q}(\boldsymbol{r}(t_{listen} + \tau)) & \cdots & \boldsymbol{q}(\boldsymbol{r}(t_{train}))
    \end{array} \right].
\end{align}
The target data matrix, $\textbf{Y}$, is constructed in a similar manner,
\begin{align}
    \textbf{Y} = \left[ \begin{array}{cccc}
        \boldsymbol{u}(t_{listen}) & \boldsymbol{u}(t_{listen} + \tau) & \cdots & \boldsymbol{u}(t_{train})
    \end{array} \right].
\end{align}
Finally $\textbf{W}_{out}$ is calculated as,
\begin{align}
    \textbf{W}_{out} = \textbf{Y} \textbf{X}^{T} \left( \textbf{X} \textbf{X}^{T} + \beta \, \textbf{I} \right)^{-1},\label{WoutRegression}
\end{align}
here $\beta$ is the regularization parameter, its role is to discourage overfitting and penalise large elements of $\textbf{W}_{out}$ from occurring. $\textbf{I}$ is the identity matrix of the appropriate dimension.

\subsection{\label{sec:Predict}Predicting Stage}

\begin{figure}
    \centering
    \includegraphics[width=0.48\textwidth]{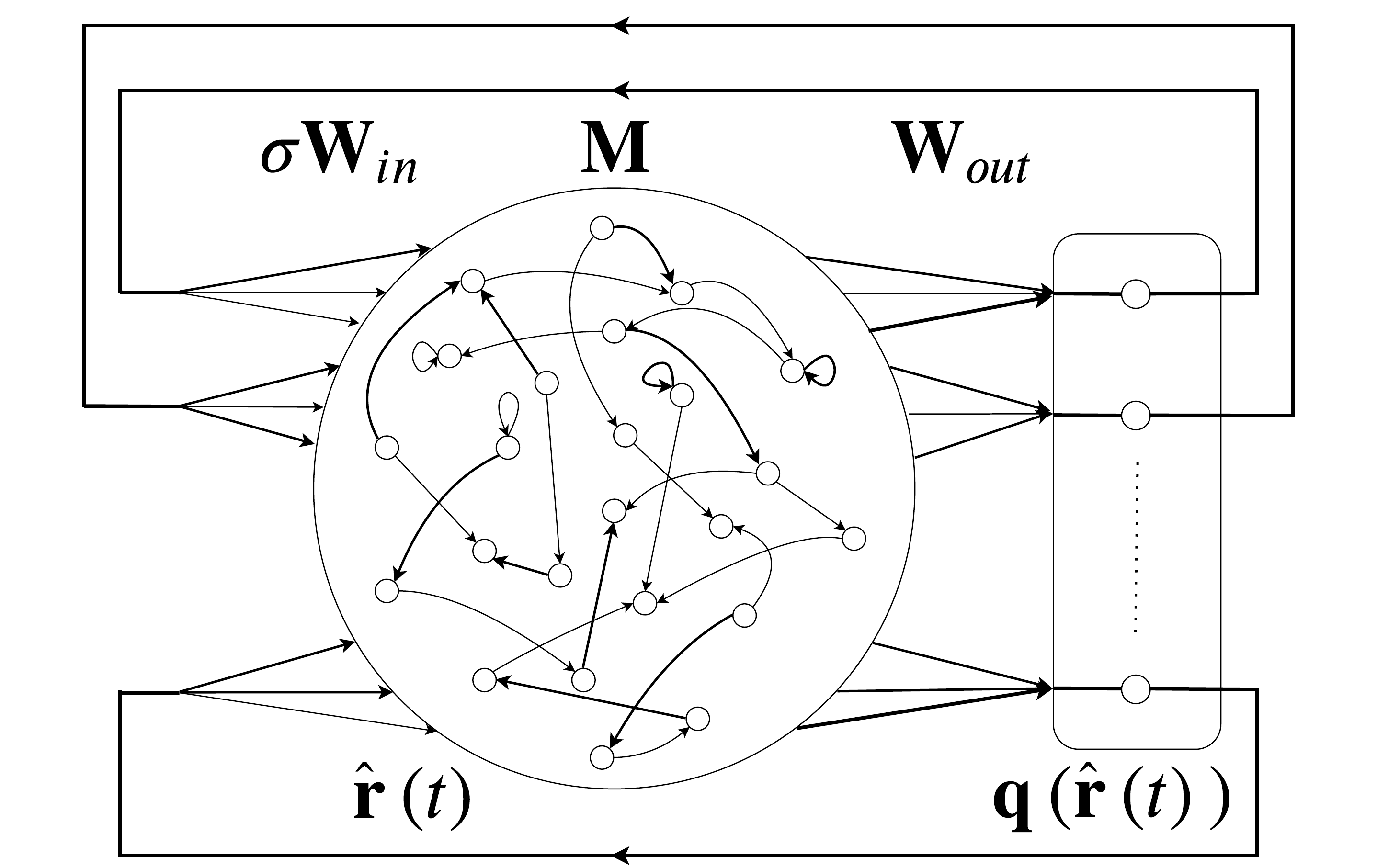}
    \caption{Predicting reservoir: The readout layer $\textbf{W}_{out} \, \boldsymbol{
    q}(\hat{\boldsymbol{r}}(t))$ replaces the external input to the reservoir.}
    \label{fig:PredResFig}
\end{figure}

In the predicting stage we compute solutions of the `predicting reservoir' described by the following equation,
\begin{align}
    \dot{\hat{\boldsymbol{r}}}(t) = \gamma \left[ - \hat{\boldsymbol{r}}(t) + \tanh{\left( \,\, \textbf{M} \, \hat{\boldsymbol{r}}(t) + \sigma \textbf{W}_{in}  \textbf{W}_{out} \, \boldsymbol{
    q}(\hat{\boldsymbol{r}}(t)) \,\, \right)} \right], \label{PredRes}
\end{align}
with $\hat{\boldsymbol{r}}(0) = \boldsymbol{r}(t_{train})$. This setup is illustrated in Fig.\,\ref{fig:PredResFig}.

If the training was successful, the readout from the reservoir, $\textbf{W}_{out} \, \boldsymbol{q}(\hat{\boldsymbol{r}}(t))$, should be an approximation of the original input which we denote as, $\hat{\boldsymbol{u}}(t) \approx \boldsymbol{u}(t) $.

In our numerical experiments we set, $N = 1000$ and the reservoir state is initialised at the beginning of the listening stage as $\boldsymbol{r}\left( 0 \right) = \left( 0, 0, \ldots, 0 \right)^{T} = \boldsymbol{0}^{T}$ for all simulations. Here $T$ denotes the transpose operation. $\textbf{M}$ is constructed as a random matrix of sparse Erd\"{o}s-Renyi connectivity with a specific spectral radius, $\rho$. To elaborate, the matrix is designed such that each element is chosen independently to be nonzero with probability $P = 0.04$ (i.e. sparsity $= 0.04$ or degree $= 40$) and these nonzero elements are chosen uniformly from $\left( -1, 1 \right)$. This random sparse matrix is rescaled such that the magnitude of its largest eigenvalue is $\rho$. For example, if $\rho$ is close to $1$, this in effect means that the input takes a long time to die out within the reservoir, which is more preferable in tasks requiring a large memory \cite{lukovsevivcius09_RCtheory}. The $\textbf{W}_{in}$ matrix is designed such that each row has only one nonzero randomly assigned element, chosen uniformly from $\left( -1, 1 \right)$. The time damping factor is kept constant at $\gamma = 10$ while $\rho$, $\sigma$ and $\beta$ are varied in order to achieve a desired learning outcome.

We remark that it is difficult to choose $\rho$, $\sigma$, $\gamma$, and $\beta$ for a specific task. To combat this, Gradient \cite{ThiedeParlitz19gradient} and Bayesian \cite{Yperman16BayesianAlg} based parameter optimisation algorithms have been developed to reduce the error in time series prediction problems. Such methods go beyond the scope of the current paper but may be useful in further studies. Instead, it is our focus to understand the mechanisms which give rise to multifunctionality in a RC. We do this by exploring the dynamics exhibited by the RC in a range of parameter values. As we will see, a multifunctional RC requires that for a given set of parameters, it can reconstruct the climate of more than one attractor. However, certain attractors may require different parameter settings to be reconstructed. In this paper we investigate the effects in performance that changing $\rho$ has in terms of the RC reconstructing a pair of attractors and therefore achieving multifunctionality. We choose to work with a random Erd\"{o}s-Renyi topology in order to provide the RC with enough dynamical flexibility to solicit multistable dynamics.

Next, we present the training technique we designed that combines data from different sources in order to construct a single output matrix that allows for the reconstruction of more than one chaotic attractor.

\subsection{\label{sec:BlendTech}Training with the `Blending Technique'}

We adapt the regression procedure from Sec.\,\ref{sec:Train} to instead use data from two input sources and the corresponding reservoir output in both training stages. From a philosophical perspective, it is necessary that the matrices $\textbf{M}$ and $\textbf{W}_{in}$ and parameters $\rho$, $\sigma$ and $\beta$ remain identical when generating both training data sets. This data is stitched together in what we call the `blending technique' with the `blending' or `weighting' parameter, $\alpha \in \left[ 0, 1 \right]$. The particular construction of the regularization matrix, $\textbf{X}$, and target data matrix, $\textbf{Y}$, used in the regression are now outlined.

First, the training data collected from the reservoir regarding each individual attractor, $\textbf{X}_{\pzs_{1}}$ and $\textbf{X}_{\pzs_{2}}$, for $\pzs_{1}$ and $\pzs_{2}$ some arbitrary attractors, are grouped together and `blended' in the following concatenation,
\begin{align}
    \textbf{X}_{C} = \left( \alpha \textbf{X}_{\pzs_{1}}, \, \left( 1 - \alpha \right) \textbf{X}_{\pzs_{2}} \right). \label{blending_technique}
\end{align}
Based on this construction, when $\alpha = 0$ or $\alpha=1$, one data set completely dominates the training. The same procedure is applied to the corresponding target data matrices in order to obtain the equivalent $\textbf{Y}_{C}$.

To avoid any biases in the concatenation step we take advantage of the memory-less based readout layer by randomly reordering each column of the matrices, $\textbf{X}_{C}$ and $\textbf{Y}_{C}$, corresponding to the input and reservoir state at a given time. These matrices are used in the ridge regression formula in Eq.\,\eqref{WoutRegression}.

\begin{figure}
    \centering
    \includegraphics[width=0.48\textwidth]{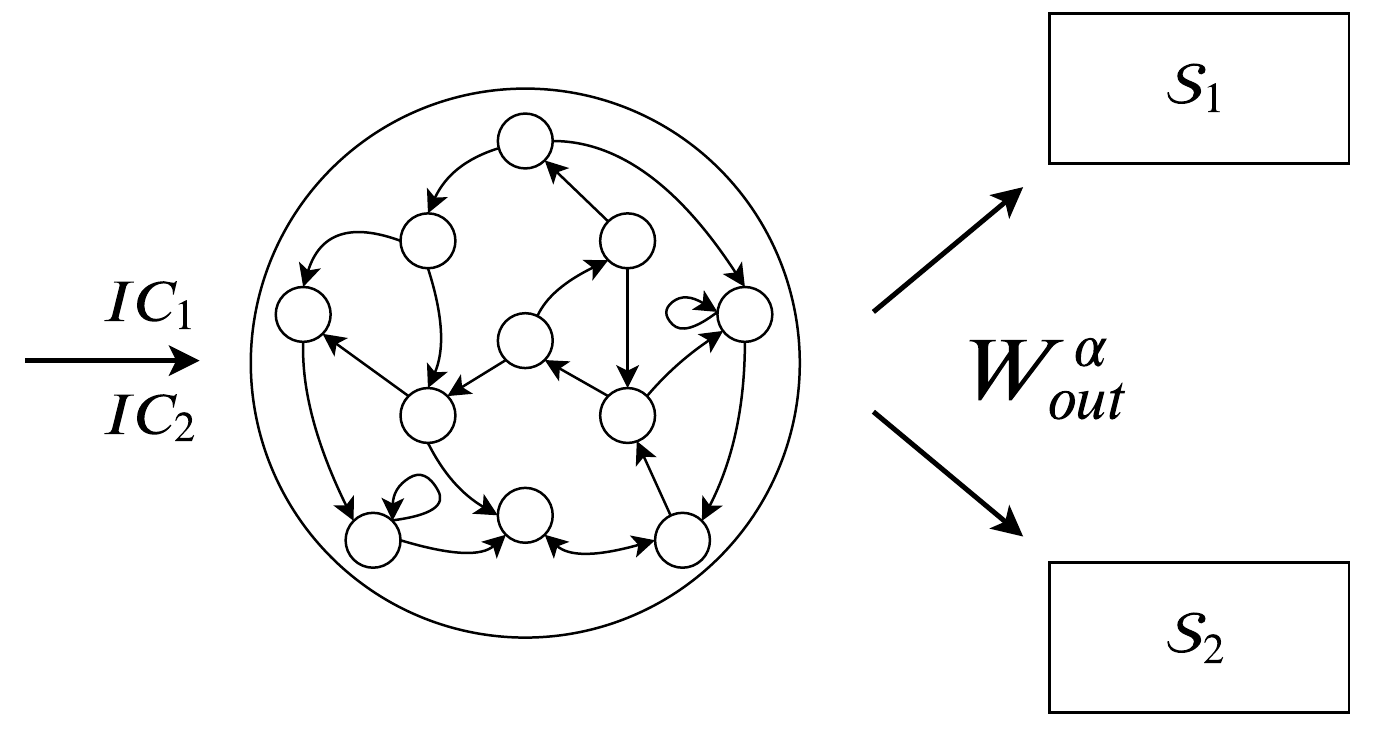}
    \caption{Illustration of a multifunctional RC}
    \label{fig:MF_Demo_Att_wout}
\end{figure}

The aim is to find an $\alpha$ that will give rise to an output matrix, $\textbf{W}^{\alpha}_{out}$, which, depending on the initial condition (IC), allows the RC to reconstruct one or the other attractor. We provide a schematic of the desired outcome in Fig.\,\ref{fig:MF_Demo_Att_wout}. 

We remark that recently a RC was used in a chaotic source separation problem where it was trained using a different method of blending signals from various chaotic sources \cite{KrishnagopalOtt20separation}. In contrast to our method, \textcite{KrishnagopalOtt20separation} consider training a RC to separate a blended input formed by a linear combination of two differently weighted chaotic signals into its constituents. Once the RC is trained on this sum of mixed signals the aim is to suppress one of the signals and continue to predict the evolution of the other chaotic time series. This work differs to the results of the current paper as we aim to train a single RC on a weighted and randomly blended training data set from two different chaotic attractors and predict the evolution of either chaotic attractor based on a given IC.

Related to the concept of multifunctionality in artificial neural networks is the notion of `systems within systems'. In this approach, distinct sub-networks are assigned particular duties within a larger network to collectively perform different behaviours. For example, the Modular Selection and Identification Control technique proposed by \textcite{WolpertKawato98} focuses on a `modular neural network' setup whereby a given trained network is singled out amongst others and brought into operation by a responsibility signal in order to generate the specific behaviour required from it. Furthermore, it was demonstrated by \textcite{TaniItoSugita04} that multiple temporal patterns can be learned using the Recurrent Neural Network with Parametric Biases setup. Here the parametric biases act as bifurcation parameters which change the dynamical regime of the network in order to generate a specific behaviour. In contrast, our approach is not modular nor requires a conscious change in parameters in order to reconstruct different attractors using the same network. When the training is successful, the multifunctional RC setup presented in this paper operates on a global scale where the artificial neurons are specifically organised in the network to perform more than one task based on a given IC.

\section{\label{sec:MFinDS}Multifunctionality and Dynamical Systems}

The characterisation of neuronal behaviour regularly invokes the language of dynamical systems theory \cite{rabinovich06_neurodyn,briggman08multifunctional}. Conceptualising certain traits of neurons from this perspective can act as a bridge between biological and artificial neural networks where advancements in one can contribute to the other.

Considering the claim made in Section \ref{sec:Intro}, that a multifunctional neural network in principle resembles a system with a coexistence of attractors, and that a RC can be trained to reconstruct the climate of a chaotic attractor, we pose the following: can a RC be trained to permit a coexistence of reconstructed attractors? Such a RC can be said to, in the spirit of our claim, express multifunctionality.

To demonstrate this we consider the following tasks which require multifunctionality by training a RC to reconstruct a coexistence of chaotic attractors from:
\begin{itemize}
    \item Case I: a multistable system.
    \item Case II: a system with different parameter settings.
    \item Case III: two different systems entirely.
\end{itemize}

We remark that such tasks closely resemble the chaotic and highly variable behaviour expressed by certain multifunctional neural networks as observed in nature \cite{Mpitsos86MFNN,Popescu02MFNN}. Furthermore, coexisting chaotic attractors are also found to occur in some low-dimensional models of neuronal systems \cite{HLu02MSCANN,Bao17HfNN}.

In order to provide an adequate testing ground we require input data from a system that allows for the generation of multiple coexisting chaotic attractors under a wide range of parameters. An example of such a system was presented in \textcite{Guan14_3DSys_ManyAttr} and described by the following set of equations,
\begin{align}
    \begin{array}{ccl}
        \dot{x}_{1}(t) & = & a \, x_{1}(t) - x_{2}(t) \, x_{3}(t) - x_{2}(t) + d, \\
        \dot{x}_{2}(t) & = & - b \, x_{2}(t) + x_{1}(t) \, x_{3}(t), \\
        \dot{x}_{3}(t) & = & - c \, x_{3}(t) + x_{1}(t) \, x_{2}(t).
    \end{array}\label{3DSys}
\end{align}
Here, $\boldsymbol{x}(t) = \left( x_{1}(t), x_{2}(t), x_{3}(t) \right)^{T}$, defines the state of the system at a given time $t$. $a$, $b$, $c$, and $d$ are the system parameters.

\begin{figure}
    \centering
    \includegraphics[width=0.48\textwidth]{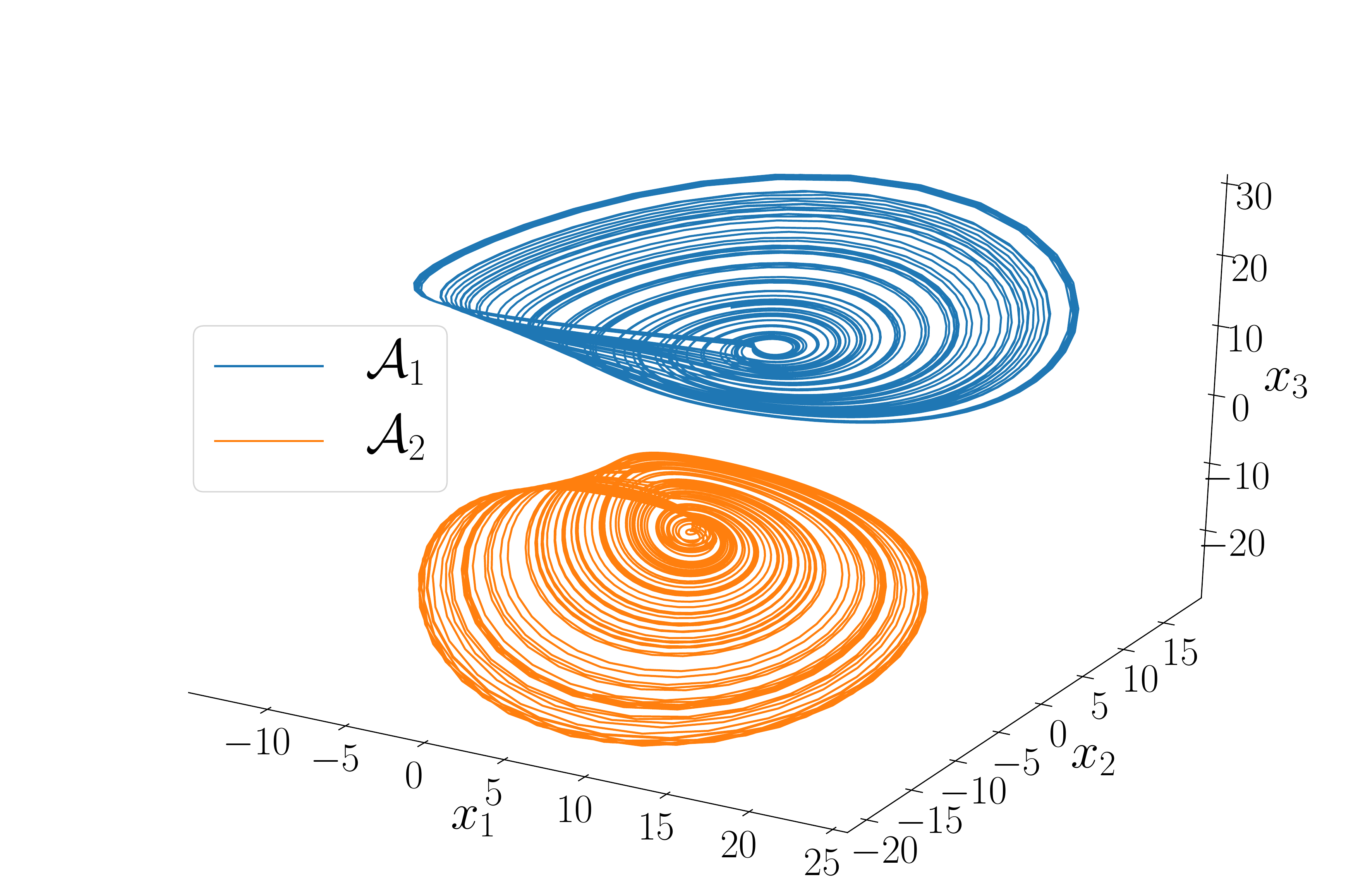}
    \caption{Coexisting attractors in the state space of Eq.\,\eqref{3DSys} for $\left( a, b, c, d \right) = \left( 5, 15, 3, 12 \right)$ and $\boldsymbol{x}^{\pzc_{1},\pzc_{2}}( 0 ) = \left( 1,1, \pm 1 \right)^{T}$.}
    \label{fig:3DSys_xyz}
\end{figure}

For example, when setting $\left( a, b, c, d \right) = \left( 5, 15, 3, 12 \right)$ there is a coexistence of two single-scroll chaotic attractors as illustrated in Fig.\,\ref{fig:3DSys_xyz}. Initialising Eq.\,\eqref{3DSys} with $\boldsymbol{x}^{\pzc_{1}}(0) = \left( 1, 1, 1 \right)^{T}$, results in the state of the system settling on the attractor $\pzc_{1}$, seen as the blue trajectory in Fig.\,\ref{fig:3DSys_xyz}. The other attractor, $\pzc_{2}$, indicated by the orange trajectory in Fig.\,\ref{fig:3DSys_xyz}, is arrived at once Eq.\,\eqref{3DSys} is initialised from $\boldsymbol{x}^{\pzc_{2}}(0) = \left( 1, 1, -1 \right)^{T}$.

Trajectories on these attractors, $\pzc_{1}$ and $\pzc_{2}$ are considered as the training data for Case I.

\begin{figure}
    \centering
    \includegraphics[width=0.48\textwidth]{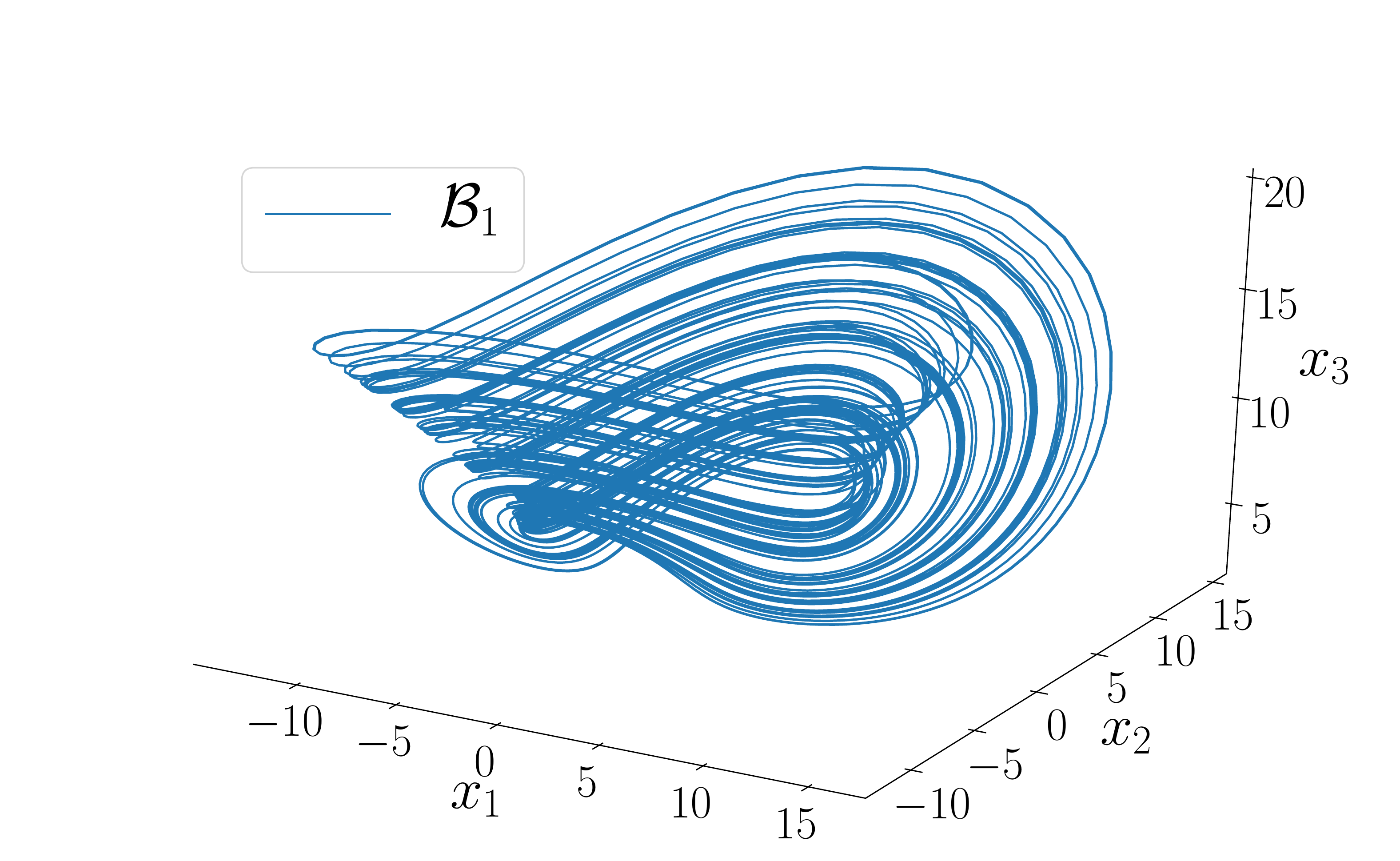}
    \caption{The attractor, $\pzb_{1}$, arrived at by computing solutions of the system Eq.\,\eqref{3DSys} initialised from $\boldsymbol{x}^{\pzb_{1}}( 0 ) = \left( 1,1,1 \right)^{T}$ with parameter values set to $\left( a, b, c, d \right) = \left( 5, 8, 2, 2 \right)$.}
    \label{fig:B1Att}
\end{figure}

In Case II we set the problem of reconstructing a double-scroll chaotic attractor in addition to the attractor $\pzc_{2}$ from Fig.\,\ref{fig:3DSys_xyz}. This particular double-scroll chaotic attractor, which we call $\pzb_{1}$, is reached by initialising Eq.\,\eqref{3DSys} with $\boldsymbol{x}^{\pzb_{1}}(0) = \left( 1,1,1 \right)^{T}$ and setting the system parameters as $\left( a, b, c, d \right) = \left( 5, 8, 2, 2 \right)$ as shown in Fig.\,\ref{fig:B1Att}. 

In Case III, we consider reconstructing $\pzc_{2}$ and the chaotic butterfly attractor, $\pzl$, generated by the Lorenz system \cite{lorenz63model}.

We remark that, the Lorenz attractor $\pzl$ and the attractors chosen from Eq.\,\eqref{3DSys} share no common region in state space. Otherwise, this adds a further level of difficulty in training a RC to distinguish which attractor a given trajectory belongs to. Alternatively, to expose further criteria needed for the RC to exhibit multifunctionality we suggest studying the effect of decreasing the distance between two disjoint attractors.

\section{\label{sec:Results}Results}

In this section we illustrate the events in which a RC was trained to exhibit multifunctionality by successfully reconstructing the coexistence of attractors as specified in Case I, II and III. We focus on establishing the optimal blending of the training data with respect to $\rho$, so to determine the regions in which multifunctionality was achieved. Following these observations we examine the circumstances in which the reconstruction of a given attractor fails and detect a number of `untrained attractors'; attractors residing within the prediction state space that were not part of the training. We then track the evolution of these attractors with respect to $\alpha$ and $\rho$ and uncover a number of `behind-the-scenes' bifurcations.

\subsection{\label{sec:MFRC}Attractor Reconstruction}

There are many means of assessing the accuracy of a predicted time series. In this work we choose to calculate, $\theta_{\pzs} ( \textbf{W}_{out}^{\pzs} )$ as the Normalised Root Mean Square Error (NRMSE) of the prediction in comparison to the target time series averaged over all state variables of a given attractor $\pzs$. $\theta_{\pzs} ( \textbf{W}_{out}^{\pzs} )_{i}$ for the $i^{th}$ state variable is calculated as, 
\begin{align}
    \theta_{\pzs} ( \textbf{W}_{out}^{\pzs} )_{i} = \frac{\sqrt{ \frac{1}{t_{predict}-t^{*}} \sum_{t=t_{predict} - t^{*}}^{t_{predict}}\left( u_{i}(t) - \hat{u}_{i}(t) \right)^{2}}}{\left| \max (u_{i}(t)) - \min (u_{i}(t)) \right|}.\label{NRMSE}
\end{align}
In our results we set $t_{predict}=600$ as the prediction end time and $t_{predict}-t^{*}$ is time that error sampling begins from. $u_{i}(t)$ and $\hat{u}_{i}(t)$ are the $i^{th}$ state variables of the target and predicted time series. The $\max \left( \cdot \right)$ and $\min \left( \cdot \right)$ functions are measures of the maximum and minimum value of the time series evaluated from $t_{predict}-t^{*}$ to $t_{predict}$. The closer $\theta_{\pzs} ( \textbf{W}_{out}^{\pzs} )$ is to $0$ the more accurate the prediction of $\pzs$. It was found empirically that for $\theta_{\pzs} ( \textbf{W}_{out}^{\pzs} ) > \delta = 0.35$ attractor reconstruction fails. Throughout our work we keep $\beta = 10^{-2}$, $\sigma = 0.2$ in Case I and III, and let $\sigma = 0.4$ in Case II.

\begin{figure}
    \centering
    \includegraphics[width=0.48\textwidth]{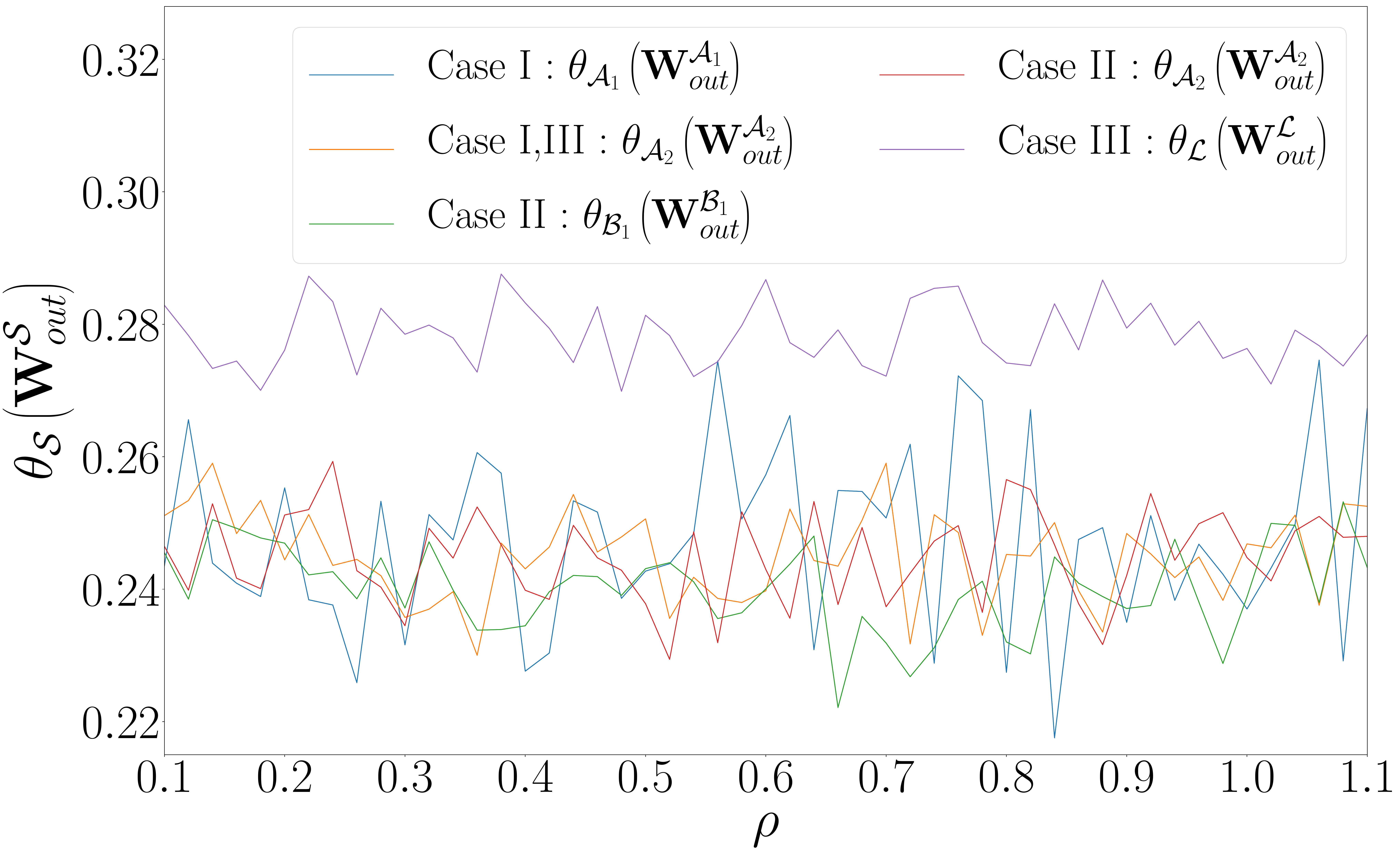}
    \caption{$\theta_{\pzs} ( \textbf{W}_{out}^{\pzs} )$\,vs.\,$\rho$ when using the task specific matrices to reconstruct the attractor $\pzs$ from Case I, II and III.}
    \label{fig:rho_NRMSE}
\end{figure}

First, we conduct an error analysis of the predicted time series when using the task specific matrices, $\textbf{W}_{out}^{\pzc_{1}}$, $\textbf{W}_{out}^{\pzc_{2}}$, $\textbf{W}_{out}^{\pzb_{1}}$, and $\textbf{W}_{out}^{\pzl}$, to determine if the attractors can be reconstructed in the usual sense. In Fig.\,\ref{fig:rho_NRMSE} we provide a picture, where having set $t^{*} = t_{train}$, the error analysis indicates attractor reconstruction was achieved for all $\rho \in \left[ 0.1, 1.1 \right]$ using each of the task specific matrices as $\theta_{\pzs} ( \textbf{W}_{out}^{\pzs} ) < \delta$ in Case I-III.

With this established we now search for values of $\alpha$ which give rise to a single readout matrix, $\textbf{W}_{out}^{\alpha}$, that allows the RC to reconstruct either of the attractors specified in Case I-III. After applying the blending technique we calculate the following error function of the particular readout matrix used to reconstruct a given attractor,
\begin{align}
    \varepsilon_{\pzs}\left( \alpha \right) = \theta_{\pzs} \left( \textbf{W}_{out}^{\alpha} \right) / \theta_{\pzs} \left( \textbf{W}_{out}^{\pzs} \right). \label{var_eps}
\end{align}
Here $\theta_{\pzs} \left( \textbf{W}_{out}^{\alpha} \right)$ is the NRMSE when using the blending technique to reconstruct an attractor $\pzs$ with a certain $\alpha$. This measure of error implies that if $\varepsilon_{\pzs}\left( \alpha \right) < 1$, the prediction of $\pzs$ was more accurate when using $\textbf{W}_{out}^{\alpha}$ over $\textbf{W}_{out}^{\pzs}$ with the opposite being said if $\varepsilon_{\pzs}\left( \alpha \right) > 1$.

\begin{figure*}
    \centering
    \begin{subfigure}{0.32\textwidth}
        \centering
        \includegraphics[width=\textwidth]{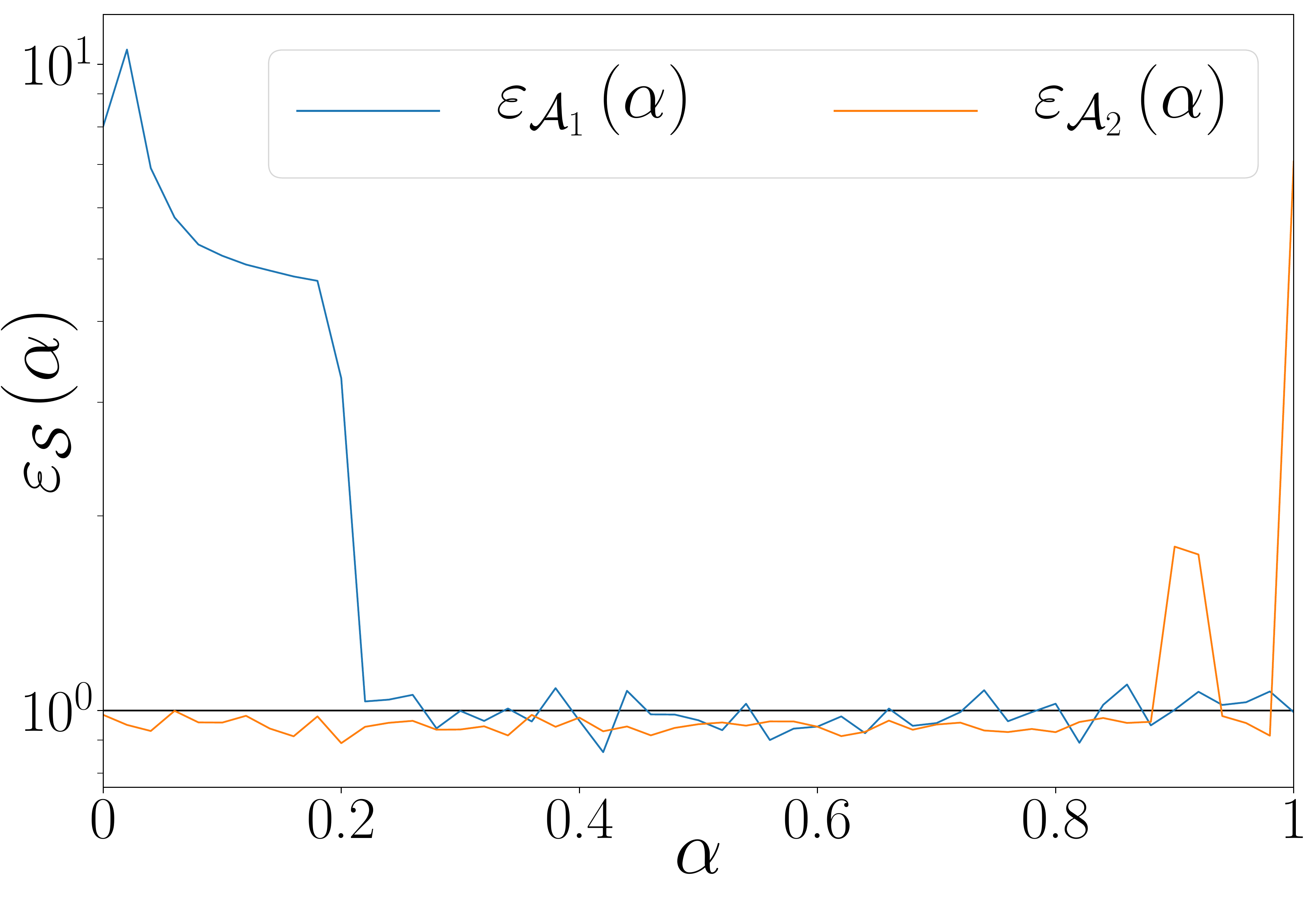}
        \caption{$\varepsilon_{\pzs}\left( \alpha \right)$\,vs.\,$\alpha$}
        \label{fig:CaseIErr}
    \end{subfigure}
    \hfill
    \begin{subfigure}{0.32\textwidth}
        \centering
        \includegraphics[width=\textwidth]{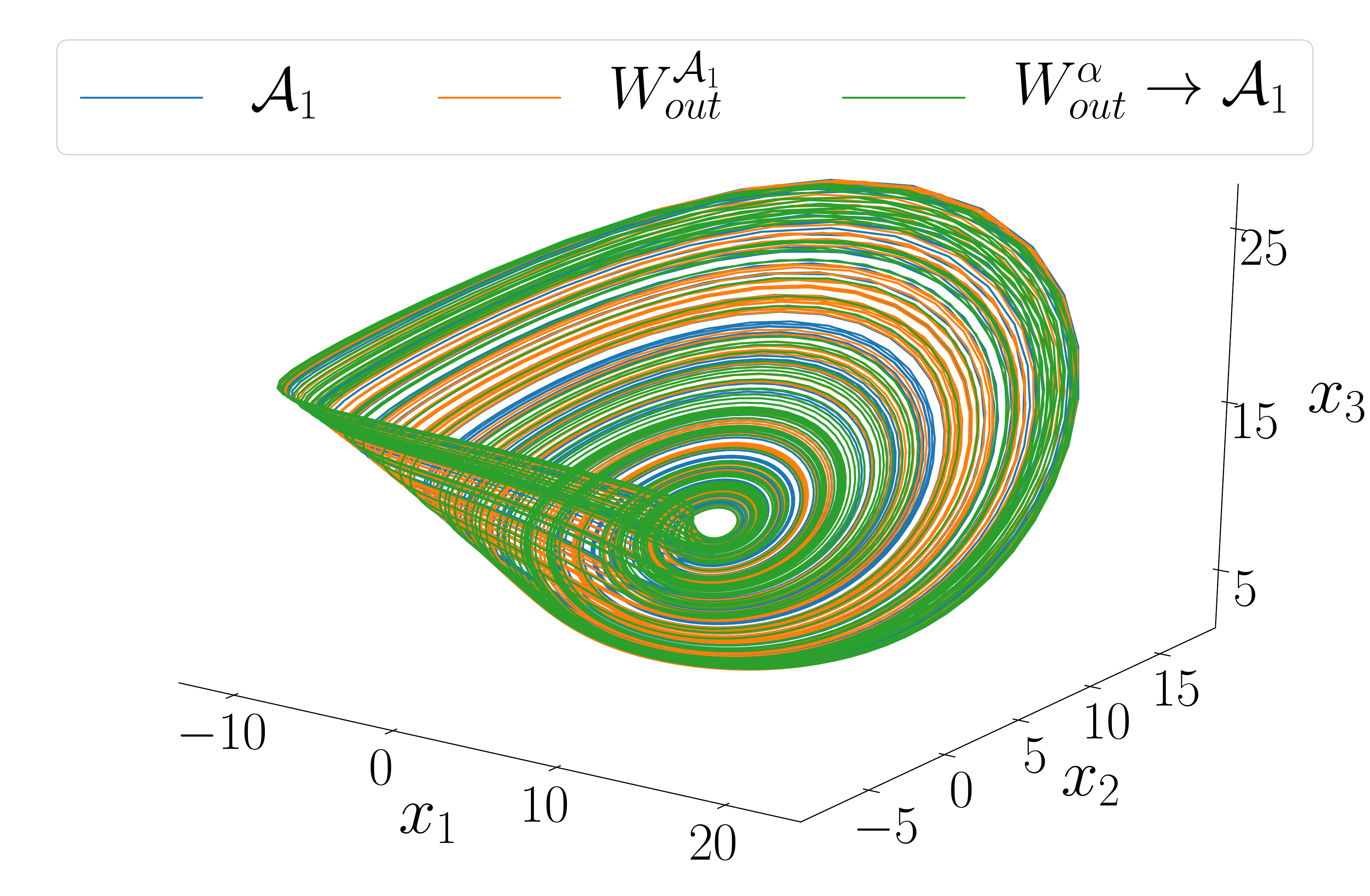}
        \caption{Reconstruction of $\pzc_{1}$}
        \label{fig:Att1_MFpred}
    \end{subfigure}
    \hfill
    \begin{subfigure}{0.32\textwidth}
        \centering
        \includegraphics[width=\textwidth]{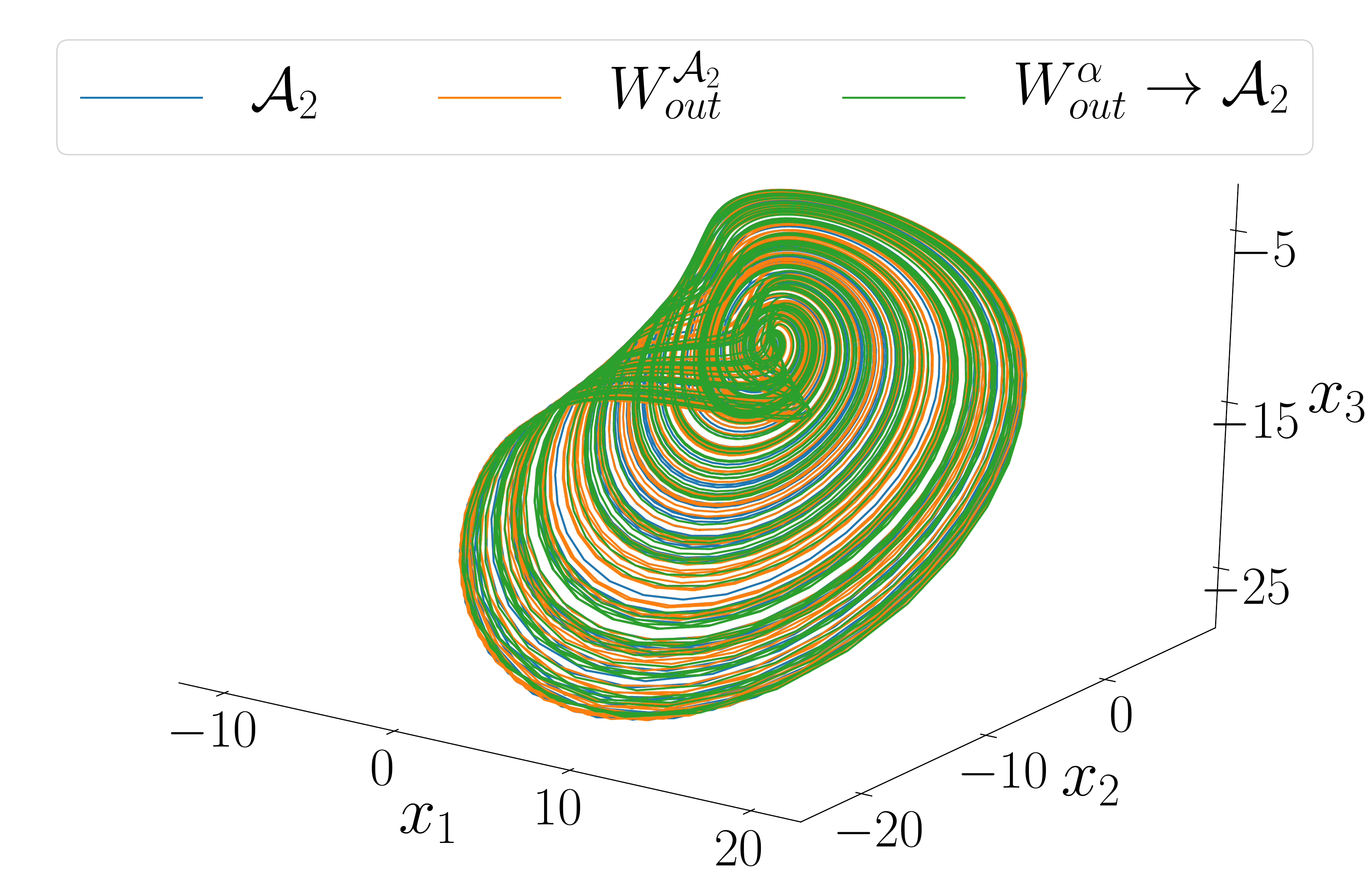}
        \caption{Reconstruction of $\pzc_{2}$}
        \label{fig:Att2_MFpred}
    \end{subfigure}
    \caption{Case I ($\rho = 0.7$) ; (a): $\varepsilon_{\pzc_{1}}$ and $\varepsilon_{\pzc_{2}}$ vs. $\alpha$. (b)-(c): Attractor reconstruction using, $\textbf{W}_{out}^{\pzc_{1}}$ and $\textbf{W}_{out}^{\pzc_{2}}$ and $\textbf{W}^{\alpha}_{out}$ for $\alpha = 0.5$.}
    \label{fig:CaseI_Err_MFpred}
\end{figure*}

Starting with the task of reconstructing attractors from a multistable system in Case I, we set $\rho = 0.7$ with the resulting $\varepsilon_{\pzs}\left( \alpha \right)$\,vs.\,$\alpha$ plot shown in Fig.\,\ref{fig:CaseIErr}.

Here we see that on one hand as $\alpha$ is increased from $0$, there is large decrease in $\varepsilon_{\pzc_{1}}$ and stays relatively close to $1$ for $\alpha \gtrsim 0.22$, while on the other hand, $\varepsilon_{\pzc_{2}}$ remains relatively near $1$ for $\alpha \lesssim 0.88$, and then grows as $\alpha$ is increased thereafter. From this we deduce that for $ 0.22 \lesssim \alpha \lesssim 0.88 $ multifunctionality is expressed by the RC. We illustrate in Figs.\,\ref{fig:Att1_MFpred}-\ref{fig:Att2_MFpred} a successful implementation of the blending technique for $\alpha = 0.5$. The predicted trajectory on $\pzc_{1}$ and $\pzc_{2}$ when using the task specific matrices, $\textbf{W}_{out}^{\pzc_{1}}$ and $\textbf{W}_{out}^{\pzc_{2}}$, are plotted in orange and the predictions using the multifunctional output matrix, $\textbf{W}^{\alpha}_{out}$, are plotted in green. A trajectory on the actual attractors from Eq.\,(\ref{3DSys}) in each figure is plotted in blue.

\begin{figure*}
    \centering
    \begin{subfigure}{0.48\textwidth}
        \centering
        \includegraphics[width=\textwidth]{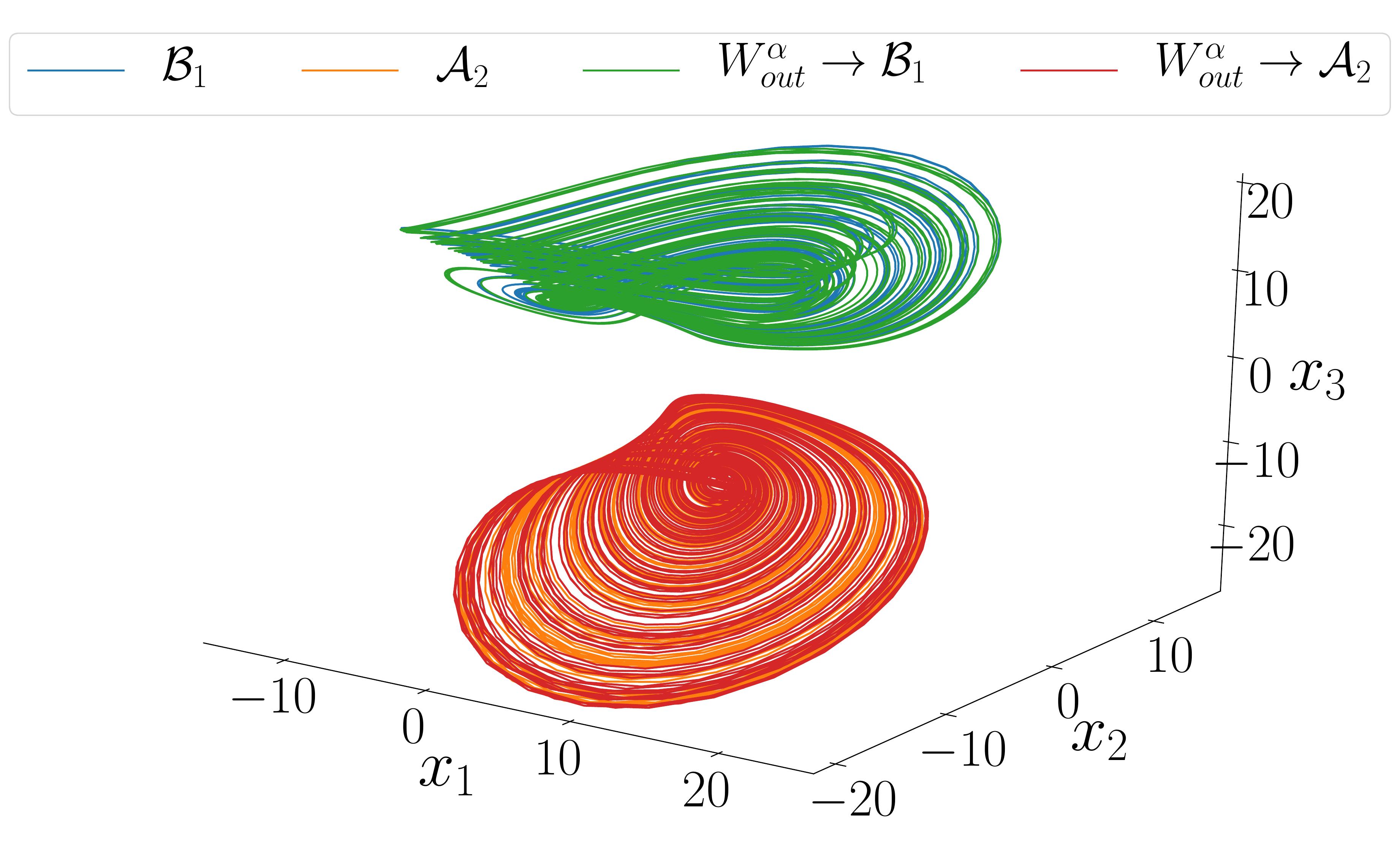}
        \caption{Case II: Reconstructing attractors $\pzb_{1}$ and $\pzc_{2}$.}
        \label{fig:3D_DiffSys_xyz}
    \end{subfigure}
    \hfill
    \begin{subfigure}{0.48\textwidth}
        \centering
        \includegraphics[width=\textwidth]{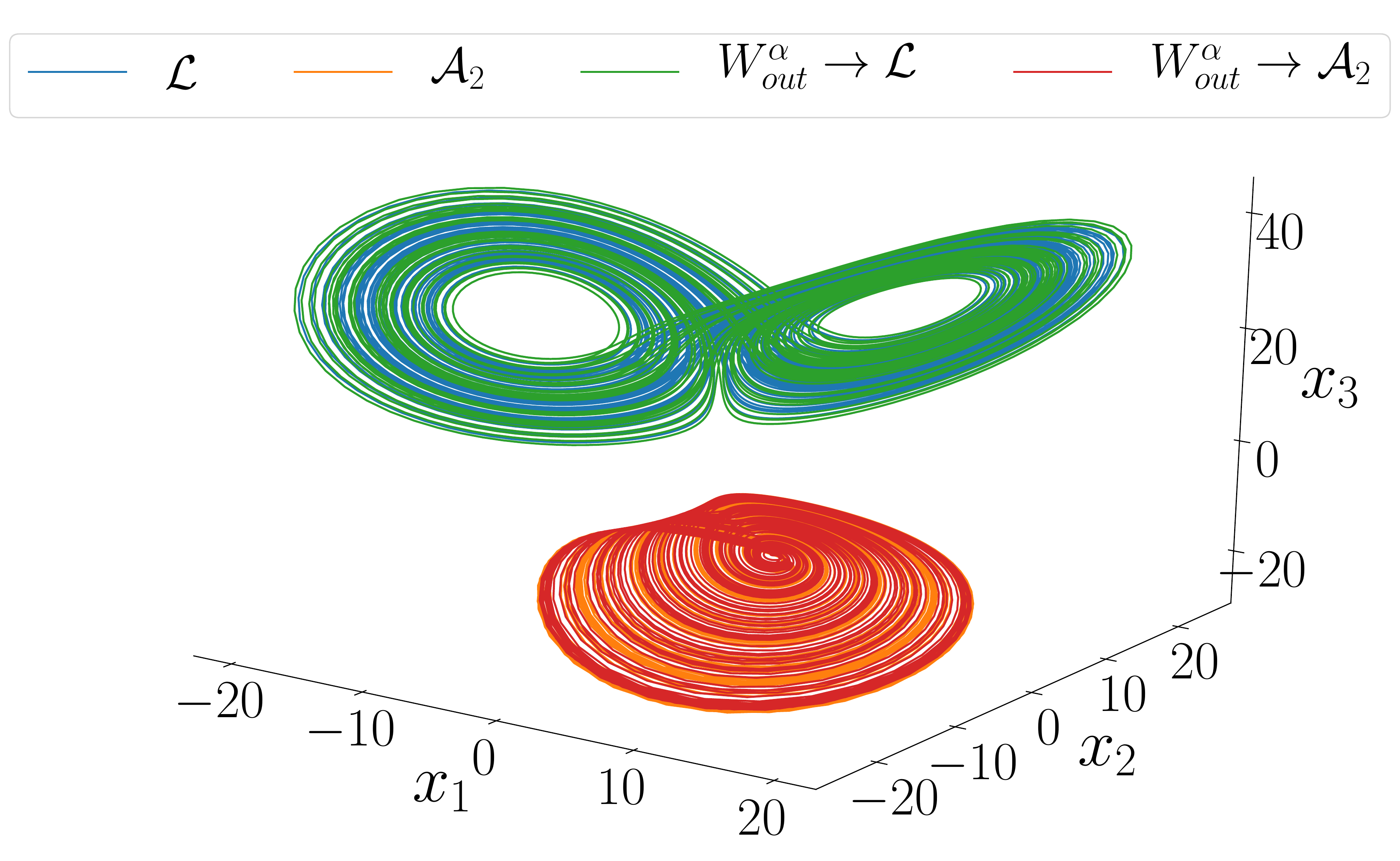}
        \caption{Case III: Reconstructing the chaotic Lorenz attractor $\pzl$ and $\pzc_{2}$.}
        \label{fig:3D_CaseIII}
    \end{subfigure}
    \caption{Illustration of attractor reconstruction in; (a) Case II for $\left( \alpha, \rho \right) = \left( 0.5, 0.3 \right)$ and (b) Case III for $\left( \alpha, \rho \right) = \left( 0.65, 0.85 \right)$}
    \label{fig:CaseII_III_Err_MFpred}
\end{figure*}

We conduct a similar analysis for Case II and III with Fig.\,\ref{fig:CaseII_III_Err_MFpred} illustrating examples where the RC was successfully trained to be multifunctional. We see in Fig.\,\ref{fig:3D_DiffSys_xyz} that when training the RC to reconstruct both $\pzb_{1}$ and $\pzc_{2}$ in Case II, the desired coexistence of chaotic attractors is achieved for $\rho = 0.3$ and $\alpha = 0.5$. We find that in Case III when setting $\rho = 0.85$ the RC is successfully trained to express multifunctionality as it can reconstruct both $\pzl$ and $\pzc_{2}$ for $\alpha = 0.65$ as seen in Fig.\,\ref{fig:3D_CaseIII}.

The results illustrated in Figs.\,\ref{fig:CaseI_Err_MFpred}-\ref{fig:CaseII_III_Err_MFpred} show that a RC can be trained to exhibit multifunctionality. Furthermore, this broadens the current set of applications a RC is capable of. We show that instead of having to change parameters in a system for it to exhibit a different behaviour, one can merge separate modes of operation from various parameter choices of the system to coexist in the prediction state space of a multifunctional RC. We also demonstrate that the combination of attractors is not limited to a single system as it is possible to combine attractors from different systems to coexist. We remark that given this ability there is the prospect of designing an appropriate controller for Eq.\,\eqref{PredRes} to switch between attractors. For further reading on `inter-attractor control' see \textcite{richter02controlchaos}. 

Although we have illustrated instances in which appropriate values of $\alpha$ were found to give rise to multifunctionality, this cannot be said for all $\alpha$ values nor for other choices of $\gamma$, $\rho$, $\sigma$ and $\beta$. Moreover, relying on an error analysis alone is not sufficient enough to classify the parameter regions in which multifunctionality is achieved. 

This RC is designed such that if attractor reconstruction fails then the predicted trajectory will not blow up to infinity in a finite amount of time. However, the prediction can decay toward some other stable attractor. Furthermore, given the nature of our study, it is also possible that the RCs predicted trajectory on one chaotic attractor can switch to the other chaotic attractor.

Following this argument, in the next section we employ a means of characterising the resultant attractor that the prediction settles to and from this identify the regions in the $\left( \alpha, \, \rho \right)$-plane where multifunctionality is achieved. 

\subsection{\label{sec:MF_in_alpha_rho}Exploring multifunctionality in the $\left( \alpha, \, \rho \right)$-plane}

In this section we analyse the long term behaviour of the RC in the prediction stage (Eq.\,\eqref{PredRes}) initialised with $\hat{\boldsymbol{r}}(0)$ corresponding to $\pzc_{1}$, $\pzc_{2}$, $\pzb_{1}$ or $\pzl$ (for the appropriate Case), and trained for a given $\alpha \in \left[ 0, 1 \right]$ and $\rho \in \left[ 0.1, 1.1 \right]$.

\begin{figure*}
    \centering
    \begin{subfigure}{0.32\textwidth}
        \centering
        \includegraphics[width=\textwidth]{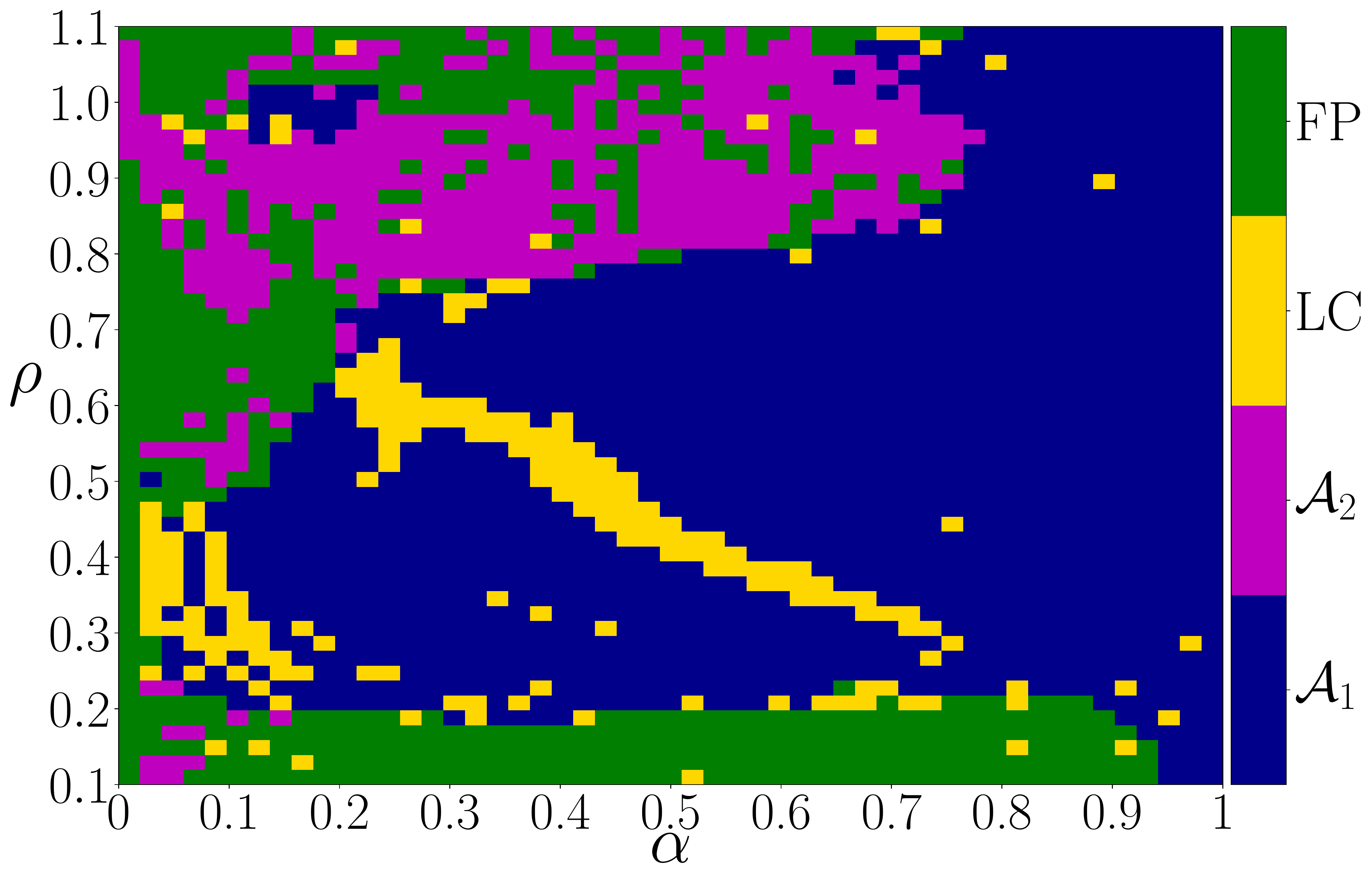}
        \caption{IC: $\pzc_{1}$}
        \label{fig:beta_2_rho_alpha_err_att1}
    \end{subfigure}
    \hfill
    \begin{subfigure}{0.32\textwidth}
        \centering
        \includegraphics[width=\textwidth]{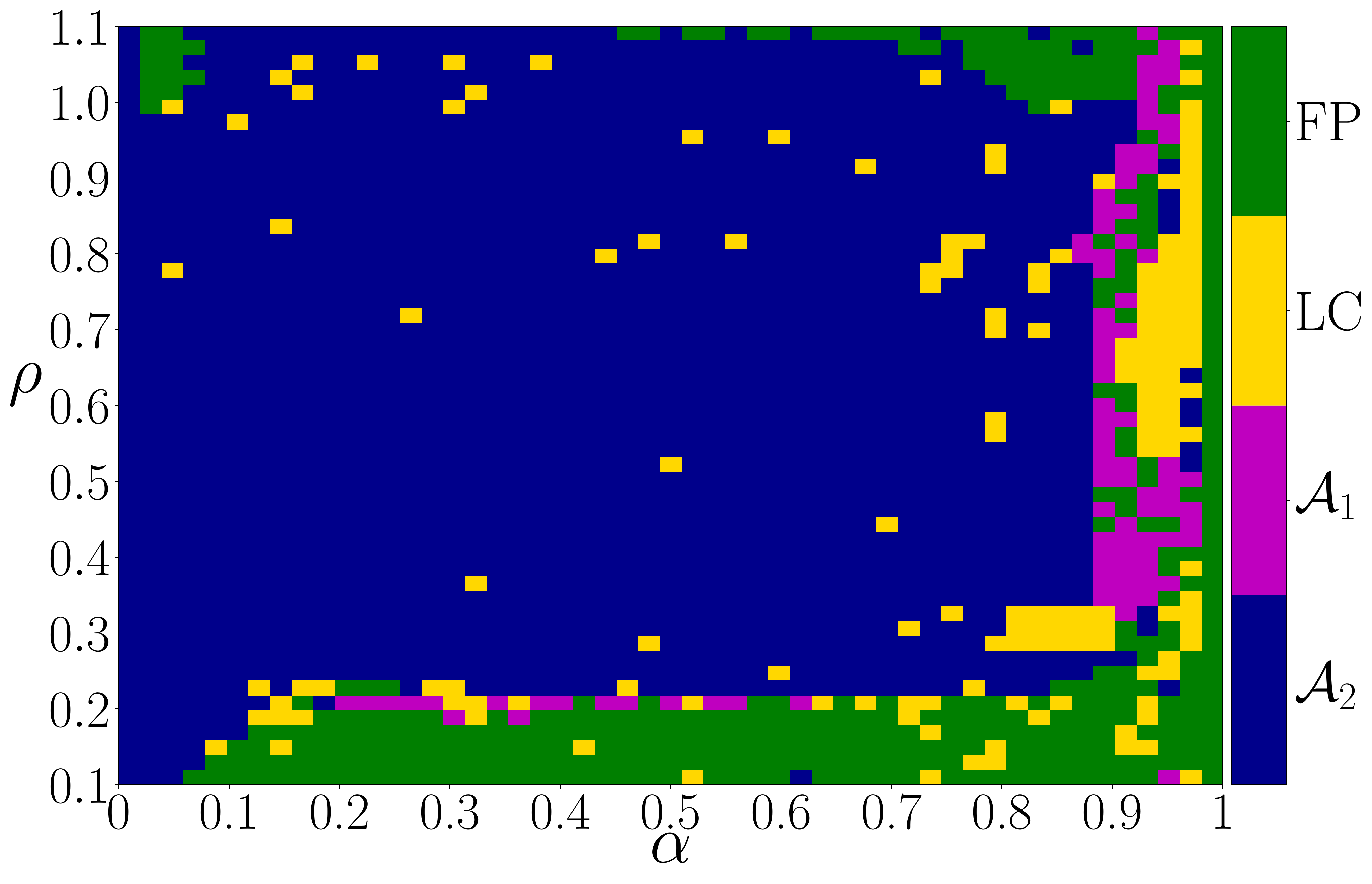}
        \caption{IC: $\pzc_{2}$}
        \label{fig:beta_2_rho_alpha_err_att2}
    \end{subfigure}
    \hfill
    \begin{subfigure}{0.32\textwidth}
        \centering
        \includegraphics[width=\textwidth]{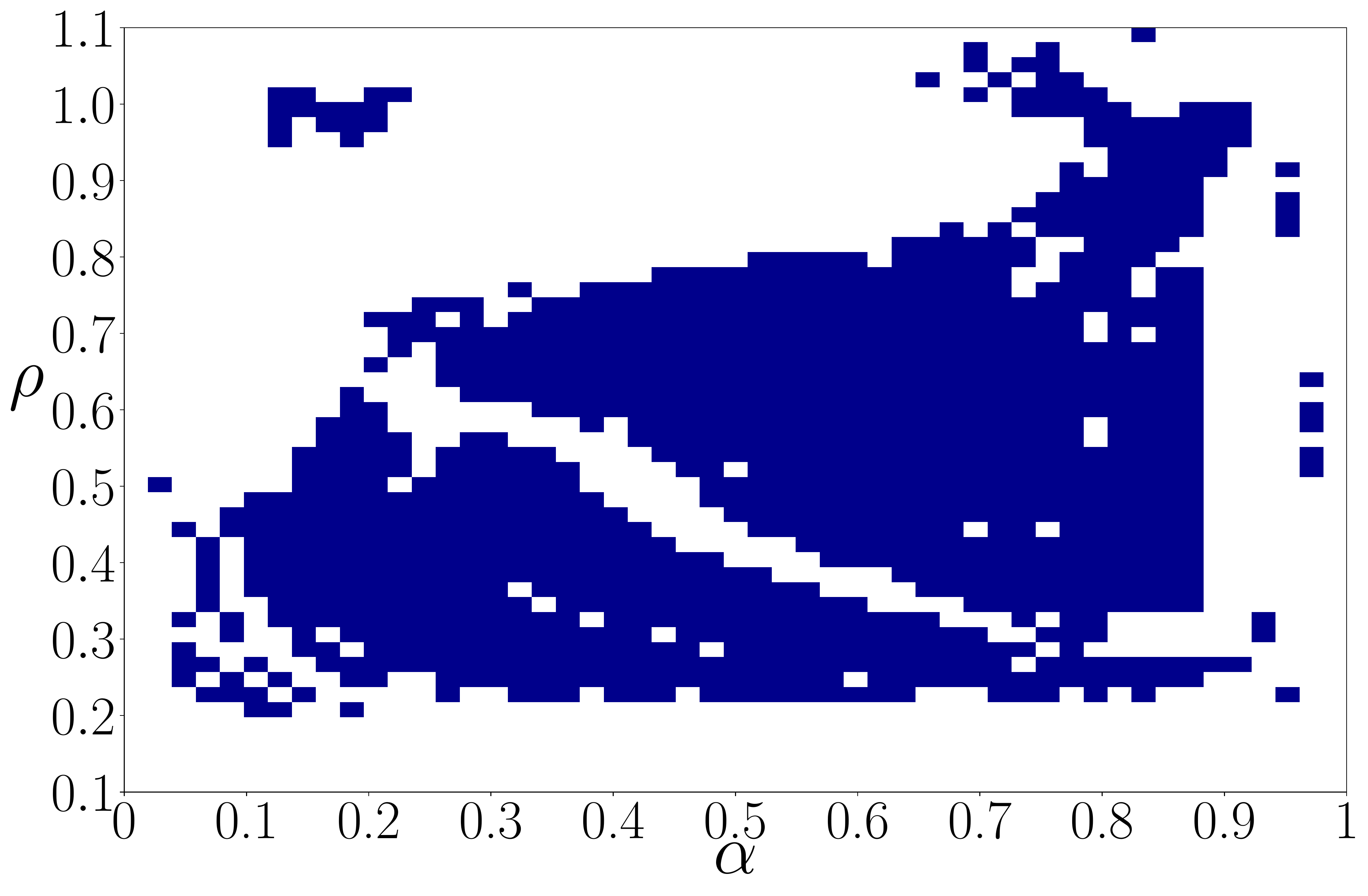}
        \caption{Regions of multifunctionality}
        \label{fig:MF_regions_Case1}
    \end{subfigure}
    \hfill
    \caption{(a)-(b):Long-term behaviour of prediction in the $\left( \alpha, \, \rho \right)$-plane for Case I. Each colour characterises the attractor the RC eventually settles to starting from a particular IC: Initial Condition. (c): Plotted in blue are the regions of multifunctionality.}
    \label{fig:beta_2_rho_alpha_err_att1_2}
\end{figure*}

We choose to assign a colour to each point in the $\left( \alpha, \, \rho \right)$-plane that characterises the prediction of a given attractor. More specifically, a point in the $\left( \alpha, \, \rho \right)$-plane is coloured:
\begin{enumerate}
    \item Yellow, if the predicted time series is periodic for $t_{predict} - 40 \leq t \leq t_{predict}$, we say the prediction has decayed to some limit cycle.
    \item Green, if the predicted time series remains constant for $t_{predict} - 10 \leq t \leq t_{predict}$, we say the prediction has decayed to some fixed point.
    \item Purple, if for $t^{*} = 40$, $\theta_{\pzs} \left( \textbf{W}_{out}^{\alpha} \right) \leq \delta$ for the predicted time series in comparison to the target time series of the other chaotic attractor, we then say that the prediction has switched from one chaotic attractor to the other.
    \item Blue if conditions 1-3 are not fulfilled and if for $t^{*} = 40$ that $\theta_{\pzs} \left( \textbf{W}_{out}^{\alpha} \right) \leq \delta$ for a given attractor $\pzs$, we then say that the climate of $\pzs$ was well reconstructed.
    \item Red if otherwise to signal that closer inspection of the prediction is needed.
\end{enumerate}
The result of this analysis for Case I is shown in Fig.\,\ref{fig:beta_2_rho_alpha_err_att1_2}. 

\begin{figure*}
    \centering
    \begin{subfigure}{0.48\textwidth}
        \centering
        \includegraphics[width=\textwidth]{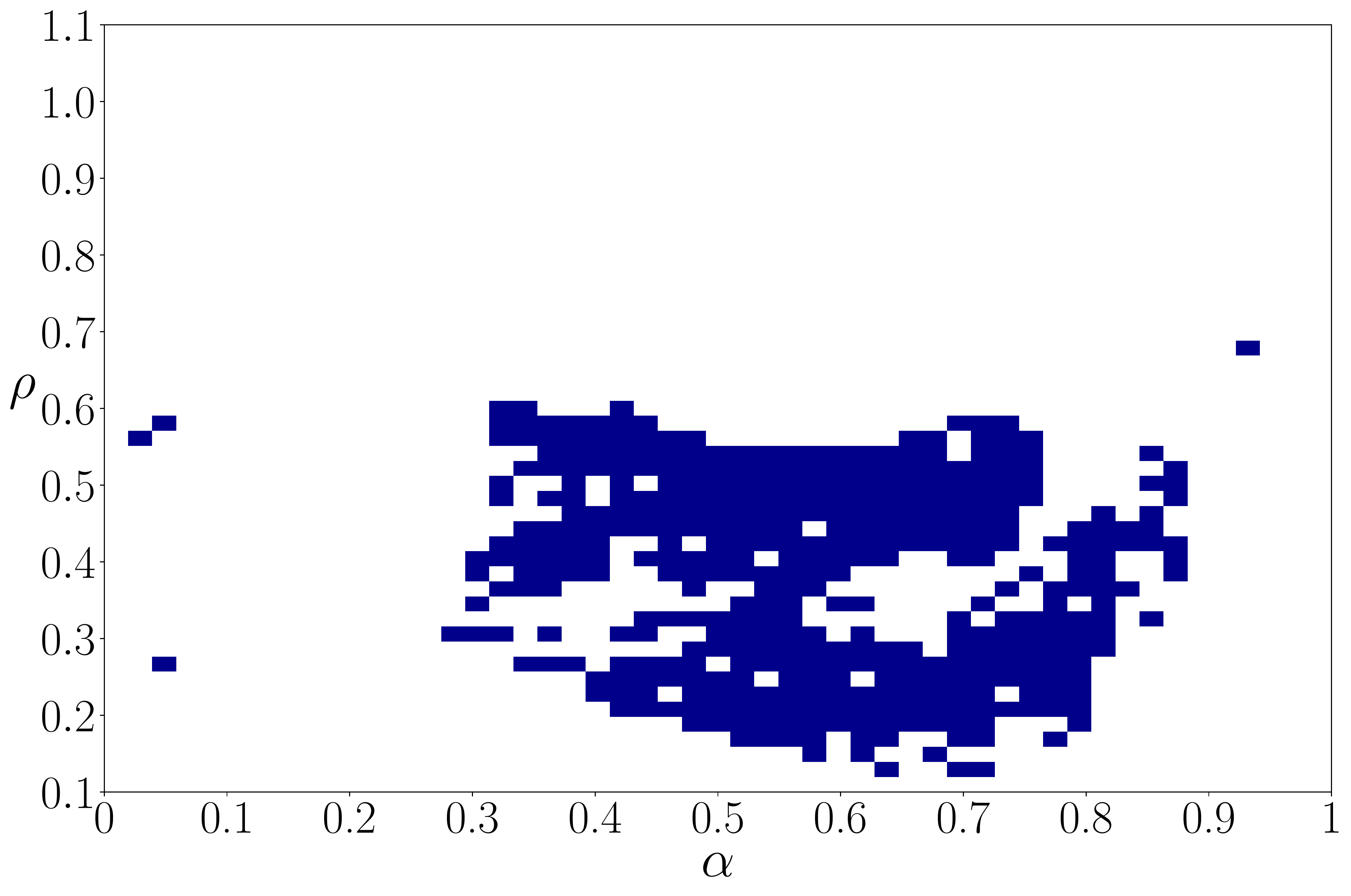}
        \caption{Case II}
        \label{fig:MF_regions_Case2}
    \end{subfigure}
    \hfill
    \begin{subfigure}{0.48\textwidth}
        \centering
        \includegraphics[width=\textwidth]{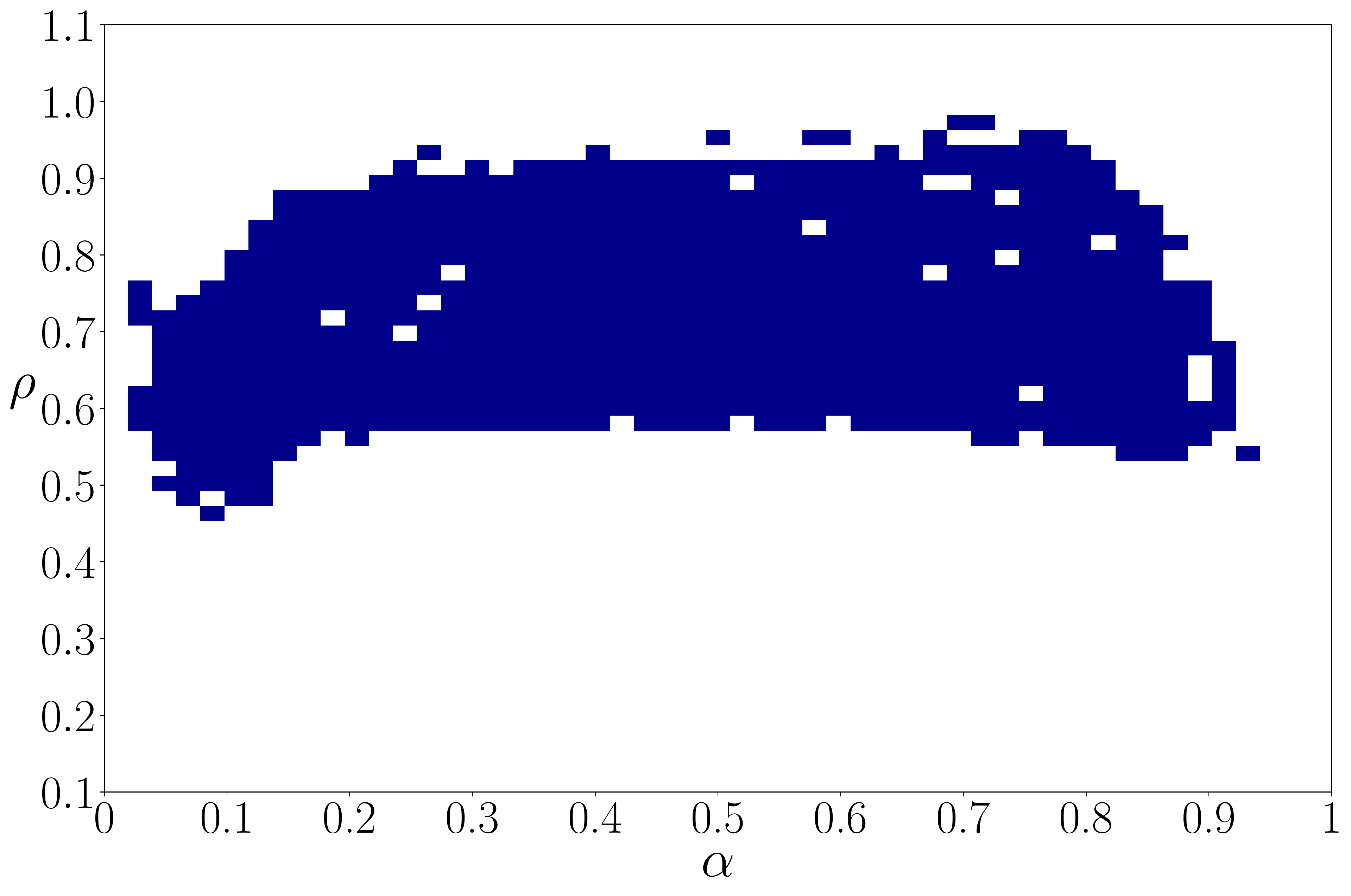}
        \caption{Case III}
        \label{fig:MF_regions_Case3}
    \end{subfigure}
    \caption{Regions of multifunctionality in the $\left( \alpha, \rho \right)$-plane for; (a) Case II and (b) Case III.}
    \label{fig:CaseII_III_MF_Regions}
\end{figure*}


As expected, we see that attractor reconstruction is impossible when initialising the RC in prediction mode with $\hat{\boldsymbol{r}}(0)$ corresponding to $\pzc_{1}$ in Fig.\,\ref{fig:beta_2_rho_alpha_err_att1} ($\pzc_{2}$ in Fig.\,\ref{fig:beta_2_rho_alpha_err_att2}) for $\alpha = 0$ ($1$). We now know that the prediction in this scenario consistently decays towards some fixed point for all values of $\rho$. However, when increasing (decreasing) $\alpha$ from $0$ ($1$) there is a more varied sequence of events for a given $\rho$. Prior to achieving attractor reconstruction the predicted trajectory on $\pzc_{1}$ ($\pzc_{2}$) at times tends towards some limit cycle or switches to $\pzc_{2}$ ($\pzc_{1}$). Fig.\,\ref{fig:beta_2_rho_alpha_err_att1} also shows that for large $\rho$, the RC favours the reconstruction of $\pzc_{2}$ as opposed to $\pzc_{1}$ where the prediction of $\pzc_{1}$ mainly switches to $\pzc_{2}$ until some critical $\alpha$ value where attractor reconstruction is achieved. Nevertheless, there is some middle ground where both attractors can be successfully reconstructed for a given pair of $\alpha$ and $\rho$, this common blue area between Figs.\,\ref{fig:beta_2_rho_alpha_err_att1}-\ref{fig:beta_2_rho_alpha_err_att2} are the regions in which we say multifunctionality was achieved. We provide a separate picture in Fig.\,\ref{fig:MF_regions_Case1} to explicitly show these regions. Through the same reasoning we show in Fig.\,\ref{fig:MF_regions_Case2} and Fig.\,\ref{fig:MF_regions_Case3} the regions in the $\left( \alpha, \rho \right)$-plane in which the RC was successfully trained to express multifunctional behaviour for both Case II and Case III.

We see a relatively large common blue region in the $\left( \alpha, \, \rho \right)$-plane for Case I in Fig.\,\ref{fig:MF_regions_Case1}. However, for Case II and III the regions of multifunctionality in Figs.\,\ref{fig:MF_regions_Case2}-\ref{fig:MF_regions_Case3} are relatively much smaller. Case II and III require a higher level of dynamical flexibility from the RC as not only do the chaotic attractors come from different settings of Eq.\,\eqref{3DSys} and different systems entirely but also vary characteristically, i.e. single-scroll and double-scroll. A particular setup of the RC may happen to favour reconstructing one flavour of attractor over the other, thus a contributing factor towards the reduced regions of multifunctionality seen here.

While in each of the explored cases the target chaotic attractors are separated in state space they do share some dynamical similarities which may be just as vital in order to achieve multifunctionality. For example, the chosen attractors in all three of the studied cases evolve along a similar timescale. We now consider the pair of chaotic attractors in Case I and investigate the relationship between the RCs capacity to exhibit multifunctionality when changing the timescale of $\pzc_{1}$ while keeping the timescale of $\pzc_{2}$ fixed. To do this, we introduce a new parameter, $\Delta^{\pzc_{1}}$ which is used to control the timescale of $\pzc_{1}$. We multiply the RHS of Eq.\,\eqref{3DSys} by $\Delta^{\pzc_{1}}$ and initialise the system with $\boldsymbol{x}^{\pzc_{1}}(0) = \left( 1, 1, 1 \right)^{T}$ to generate solutions of $\pzc_{1}$ with a modified timescale. If $\Delta^{\pzc_{1}} < 1$ the dynamics are slowed down and if $\Delta^{\pzc_{1}} > 1$ the dynamics are sped up. 
    
\begin{figure}
    \centering
    \includegraphics[width=0.49\textwidth]{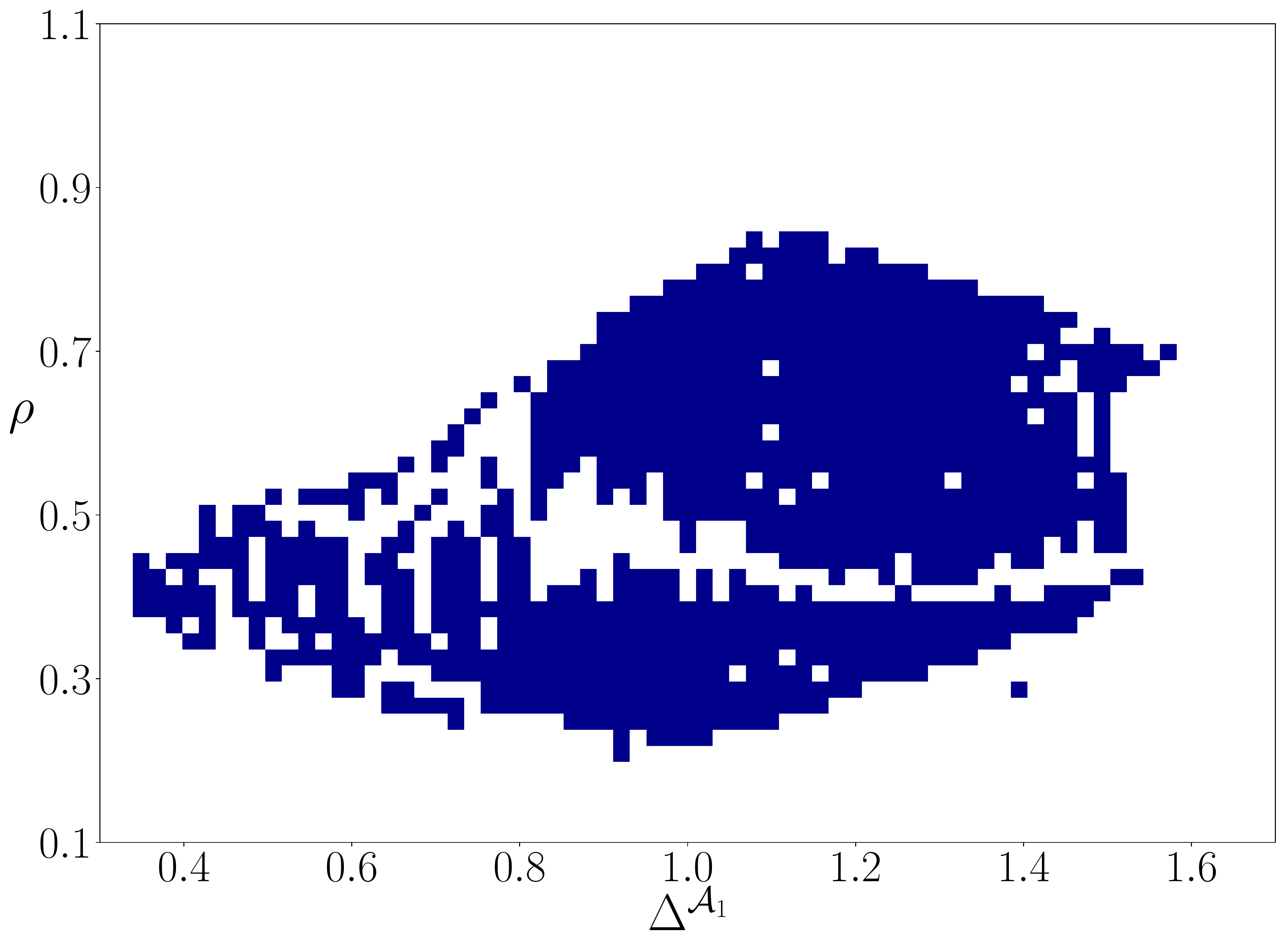}
    \caption{Regions of multifunctionality in the $( \Delta^{\pzc_{1}}, \rho )$-plane when changing the timescale of $\pzc_{1}$ in Case I with timescale parameter $\Delta^{\pzc_{1}}$.}
    \label{fig:TimescaleMF}
\end{figure}

We now investigate and identify the regions in the $( \Delta^{\pzc_{1}}, \rho )$-plane where multifunctionality is achieved. We do this by employing the previous method of characterising the long-term behaviour of the RCs prediction of either $\pzc_{1}$ or $\pzc_{2}$ when using $W_{out}^{\alpha}$ with $\alpha = 0.5$ and take the common regions where the RC can successfully reconstruct either attractor depending on the IC. The result of this is shown in Fig.\,\ref{fig:TimescaleMF}.
    
Here in Fig.\,\ref{fig:TimescaleMF} we see that if $\pzc_{1}$ evolves along a relatively longer or shorter timescale in comparison to $\pzc_{2}$, then there is a point where multifunctionality is lost for all $\rho$ values. This reveals the limits upon which multifunctionality can be achieved in this scenario by exposing the RCs dynamical capacity to facilitate the coexistence of increasingly dissimilar chaotic attractors in its prediction state space.

\begin{figure}
    \centering
    \includegraphics[width=0.48\textwidth]{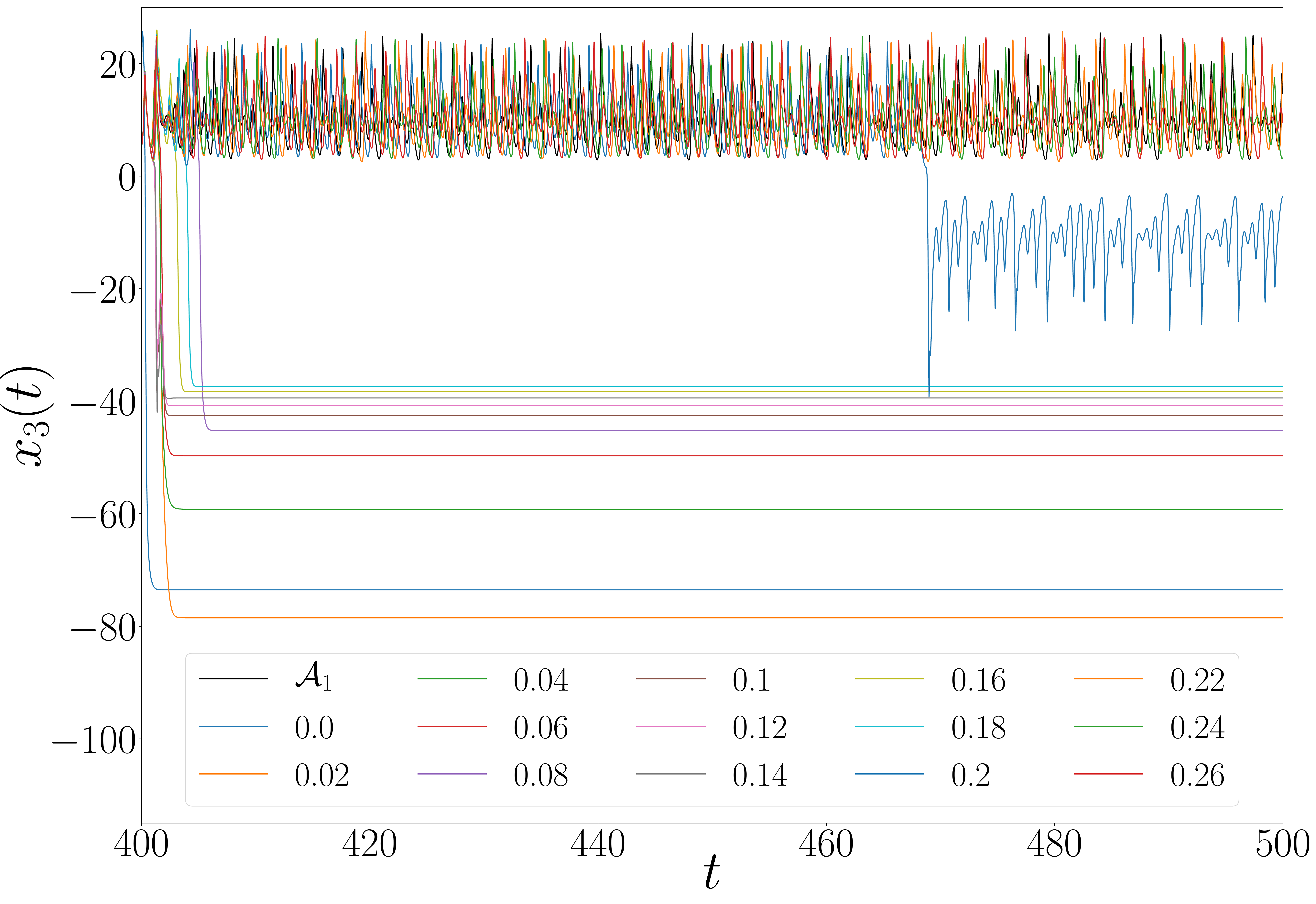}
    \caption{Case I ($\rho=0.7$, IC: $\pzc_{1}$): $x_{3}(t)$\,vs.\,$t$ for $\pzc_{1}$ coloured in black and the predicted trajectories for various values of $\alpha$ as indicated in the plot legend.}
    \label{fig:CaseI_rho_0_7_alpha_left}
\end{figure}

Now, if we focus on the prediction of $\pzc_{1}$ for $\rho = 0.7$ as $\alpha$ is increased from $0$ in Fig.\,\ref{fig:beta_2_rho_alpha_err_att1}, we see that before multifunctionality is achieved the prediction repeatedly falls to a fixed point. We plot these fixed points for various values of $\alpha$ in Fig.\,\ref{fig:CaseI_rho_0_7_alpha_left}. Also seen here is a case where the prediction switches from $\pzc_{1}$ to $\pzc_{2}$ when $\alpha = 0.2$.

We ask, what happens to these fixed points in Fig.\,\ref{fig:CaseI_rho_0_7_alpha_left} as multifunctionality is achieved? Is it the case that the successful reconstruction of $\pzc_{1}$ entirely takes the place of these fixed points in the prediction state space or, do these fixed points still exist and we are unable see them? Furthermore, are there other attractors lurking in this prediction state space that we are not immediately aware of? We delve further into these questions in the following section and highlight the important role of ICs in the prediction stage of the RC. 

\subsection{\label{sec:UntrainedAtt}Detecting Untrained Attractors}

To get a broader picture of the prediction state space for a given $\rho$ and $\alpha$ in Case I, we initialise the RC in the prediction mode with many (1000) random initial conditions (RICs) and observe the resultant trajectories.

\begin{figure}
    \centering
    \hfill
    \begin{subfigure}{0.48\textwidth}
        \centering
        \includegraphics[width=0.99\textwidth]{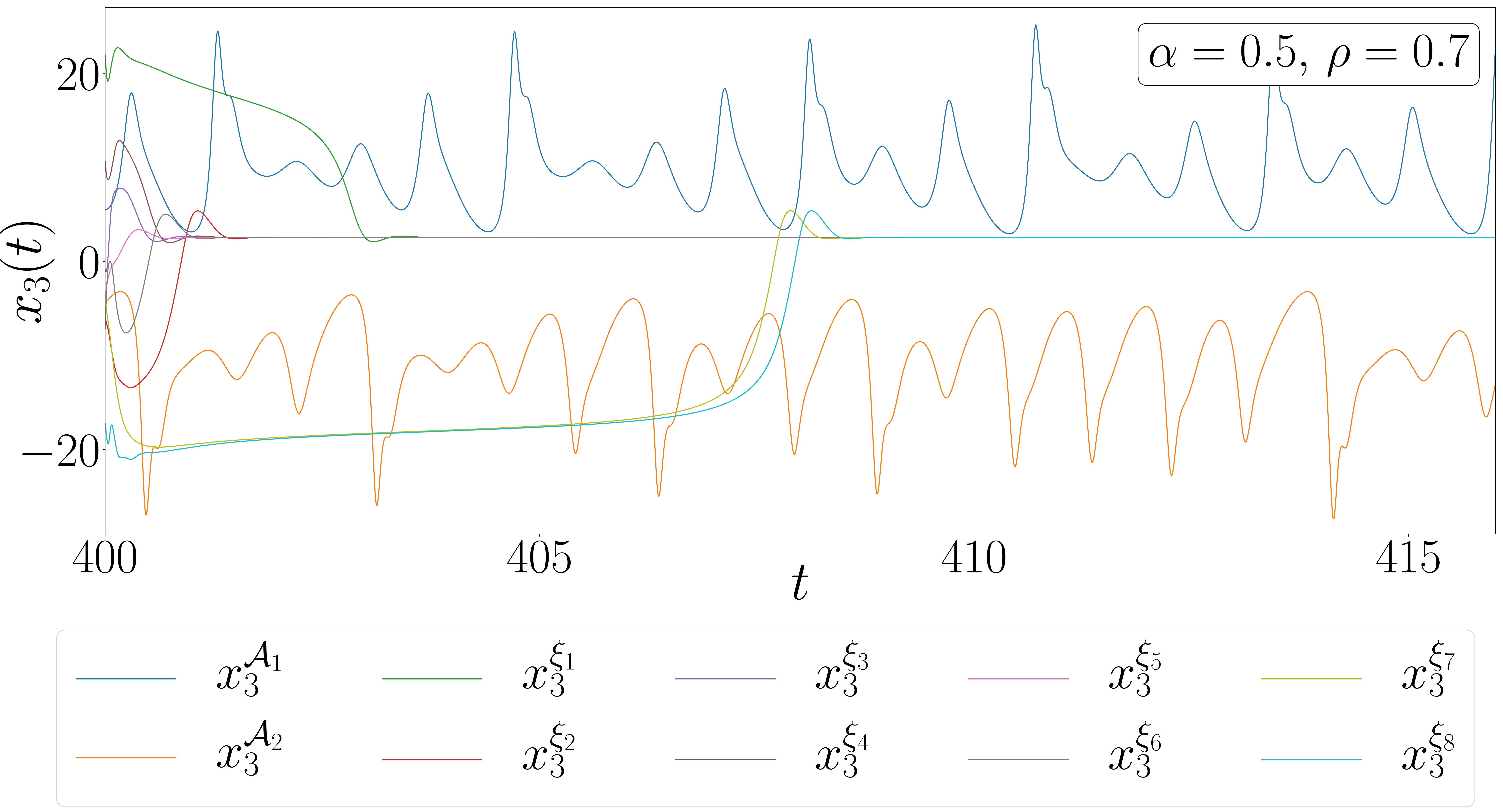}
        \label{fig:rand_FP_ev_rho_7_alpha_5}
    \end{subfigure}
    \hfill
    \vskip\baselineskip
    \begin{subfigure}{0.235\textwidth}
        \centering
        \includegraphics[width=0.99\textwidth]{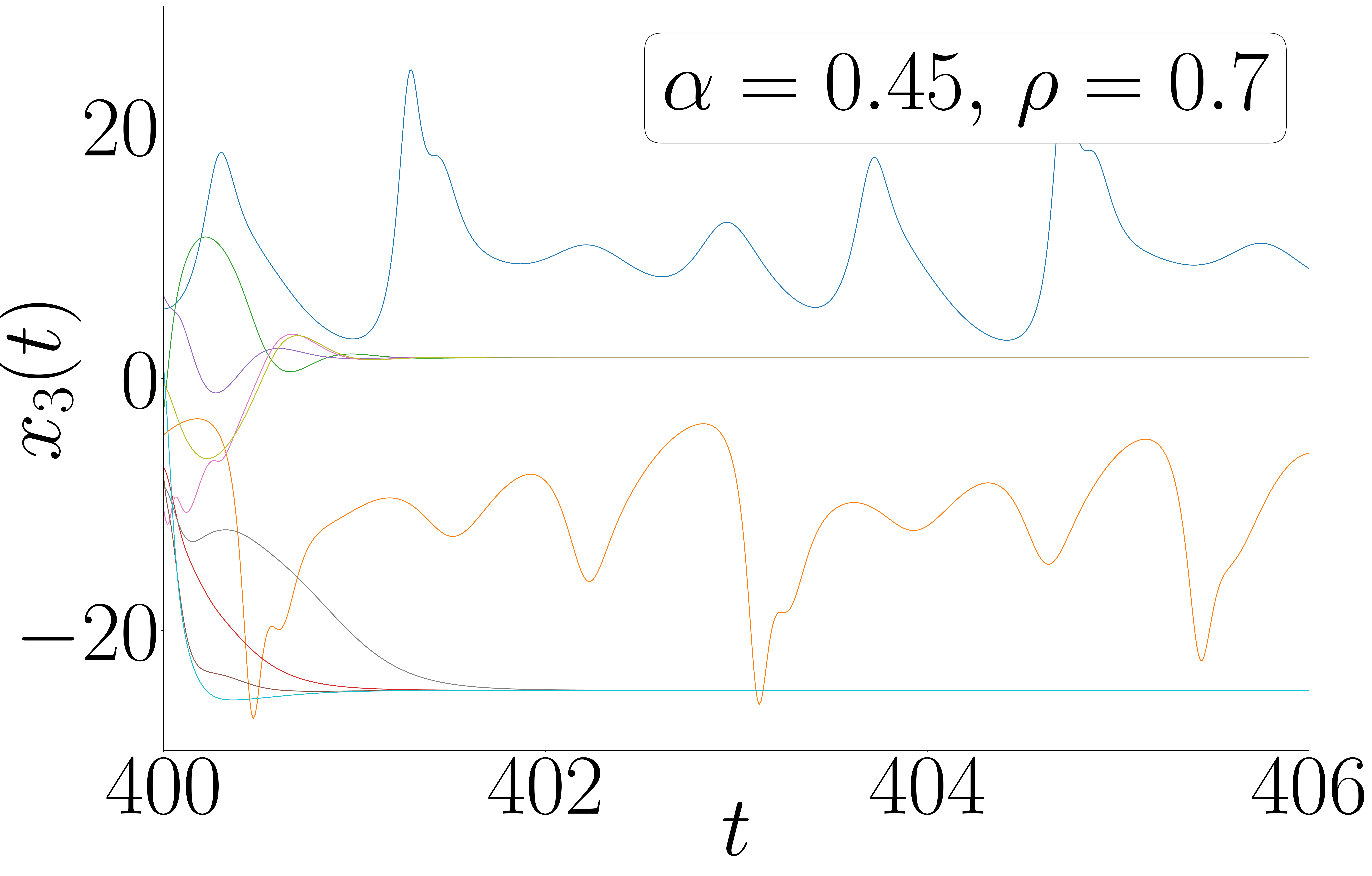}
        \label{fig:rand_FP_ev_rho_7_alpha_45}
    \end{subfigure}
    \hfill
    \begin{subfigure}{0.235\textwidth}
        \centering
        \includegraphics[width=0.99\textwidth]{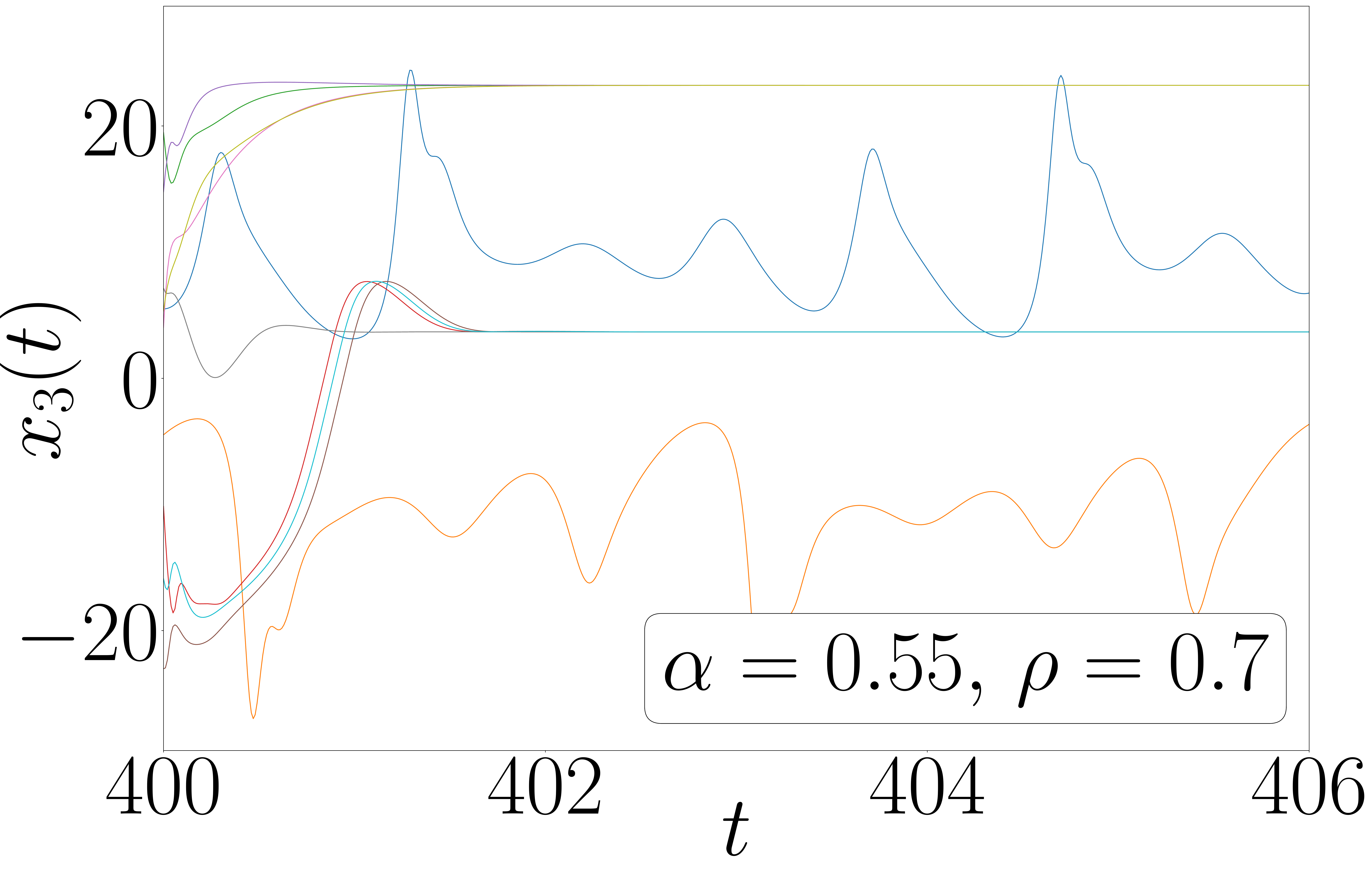}
        \label{fig:rand_FP_ev_rho_7_alpha_55}
    \end{subfigure}
    \caption{Time trace of the predicted $x_{3}$ variable, with the RC trained for a given $\alpha$ and $\rho$ as indicated above each plot, starting from the random initial condition $\boldsymbol{\xi}_{i}$ for $i=1, 2, \ldots$.}
    \label{fig:rand_FP_ev_rho_7_alpha}
\end{figure}

In Fig.\,\ref{fig:rand_FP_ev_rho_7_alpha} we set $\rho = 0.7$ and train the RC for $\alpha = 0.45$, $0.50$, and $0.55$. We plot $x_{3}^{\boldsymbol{\xi}_{i}}$ as the trajectories of the $x_{3}$ variable as predicted by the RC starting from the $i^{th}$ RIC.

For $\alpha = 0.5$ we see that there exists a stable fixed point located within the middle of both reconstructed chaotic attractors. As we begin to apply more weight to the data from $\pzc_{2}$ with $\alpha=0.45$, we see another stable fixed point appearing closer to the reconstructed $\pzc_{2}$. Similarly for $\alpha=0.55$ we find a stable fixed point closer to the reconstructed $\pzc_{1}$. So while the RC was successfully trained to reconstruct the climate of both $\pzc_{1}$ and $\pzc_{2}$ there are additional attractors populating the prediction state space that were not involved in the training, we call these the `untrained attractors'.

The results in Fig.\,\ref{fig:rand_FP_ev_rho_7_alpha} show that small changes in $\alpha$ can give rise to a bistability of untrained attractors. This suggests that there are some `behind-the-scenes' bifurcations taking place in the prediction state space. However, further discussion is needed to consider $\alpha$ as a bifurcation parameter of the RC.

While $\rho$ is a parameter of the RC itself, $\alpha$ is a parameter that is strictly involved in the training procedure. Therefore to consider $\alpha$ as a bifurcation parameter of the RC we need to assess if for a small change in $\alpha$ there is a relatively smooth change in the elements of the $\textbf{W}_{out}^{\alpha}$ matrix generated from a specific combination of $\alpha$ and $\rho$. In other words, a continuous function which maps $\left( \alpha, \, \rho \right) \rightarrow \textbf{W}_{out}^{\alpha}$ needs to exist.

\begin{figure}
    \centering
    \includegraphics[width=0.48\textwidth]{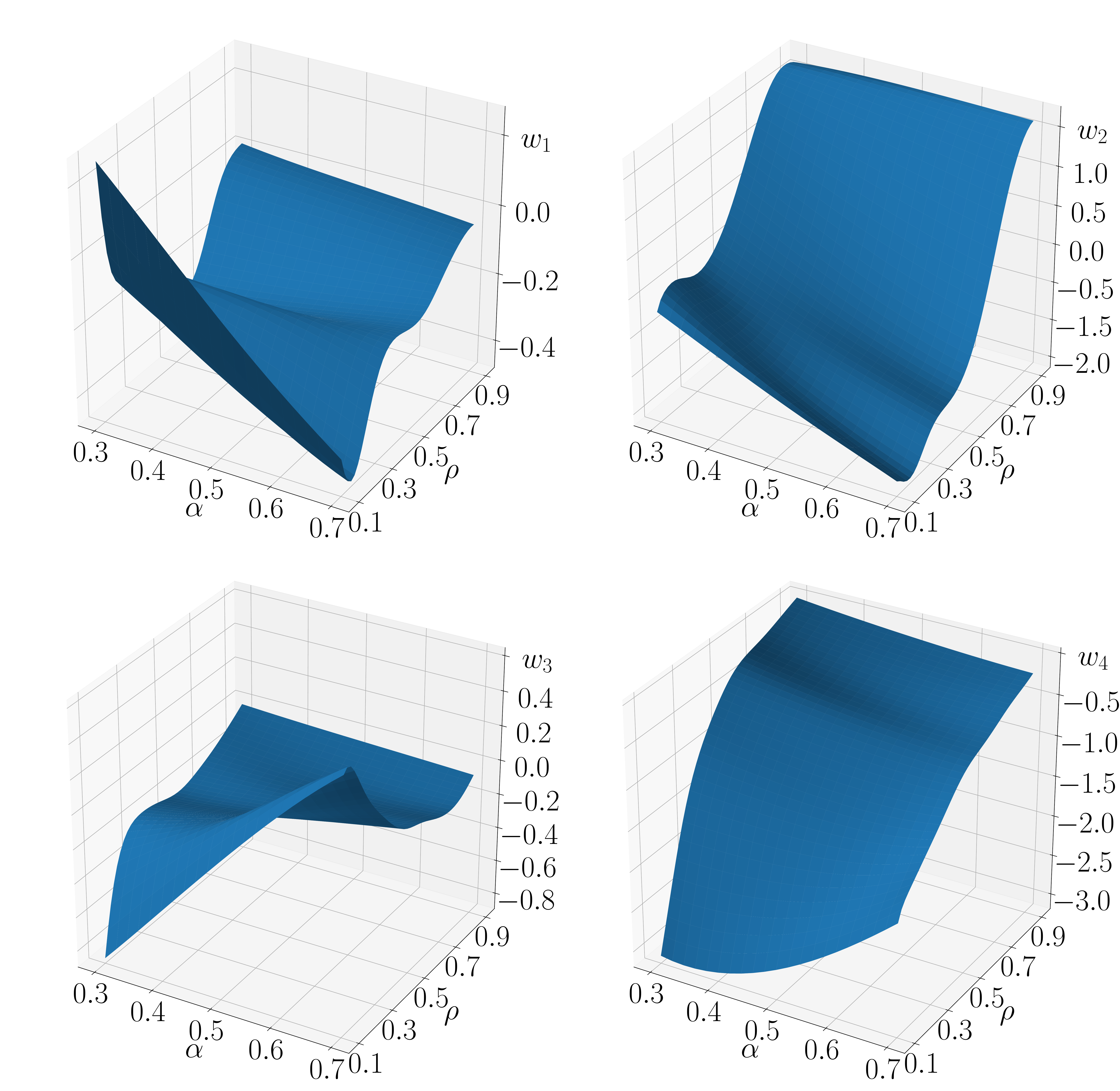}
    \caption{The evolution of four randomly chosen elements of the $\textbf{W}_{out}^{\alpha}$ matrix with respect to changes in $\alpha$ and $\rho$.}
    \label{fig:WoutSurf}
\end{figure}

To expand upon this notion we consider some randomly chosen elements of $\textbf{W}_{out}^{\alpha}$ and observe their evolution with respect to $\alpha$ and $\rho$. The result of this is shown in Fig.\,\ref{fig:WoutSurf} for four randomly chosen elements of $\textbf{W}_{out}^{\alpha}$. 

The smoothness of these surface plots indicate that a relatively small change in $\alpha$ or $\rho$ in turn gives rise to a small change in the elements of $\textbf{W}_{out}^{\alpha}$, to which we generalise, contributes to an overall change in the dynamics of the RC. From this we now consider $\alpha$ as a bifurcation parameter of the RC. We can view our training procedure as a smooth map from the hyper-parameters $\rho$ and $\alpha$ to a matrix $\textbf{W}_{out}^{\alpha}$ in the $\mathbb{R}^{D \times 2 N}$ space. Therefore by considering these hyper-parameters, $\rho$ and $\alpha$, this mapping generates a two-dimensional structure in the $\mathbb{R}^{D \times 2N}$ space.

Our study now moves towards tracking the evolution of these untrained attractors and identifying some of these `behind-the-scenes' bifurcations.

\subsection{\label{sec:BehaveUntrainedAtt}Bifurcation Analysis of Untrained Attractors}

In this section we track the evolution of these untrained attractors firstly in the $\left( \alpha, \, x_{3} \right)$-plane for a given $\rho$ and then in the $\left( \alpha, \rho \right)$-plane.

\begin{figure*}
    \centering
    \begin{subfigure}{0.48\textwidth}
        \centering
        \includegraphics[width=\textwidth]{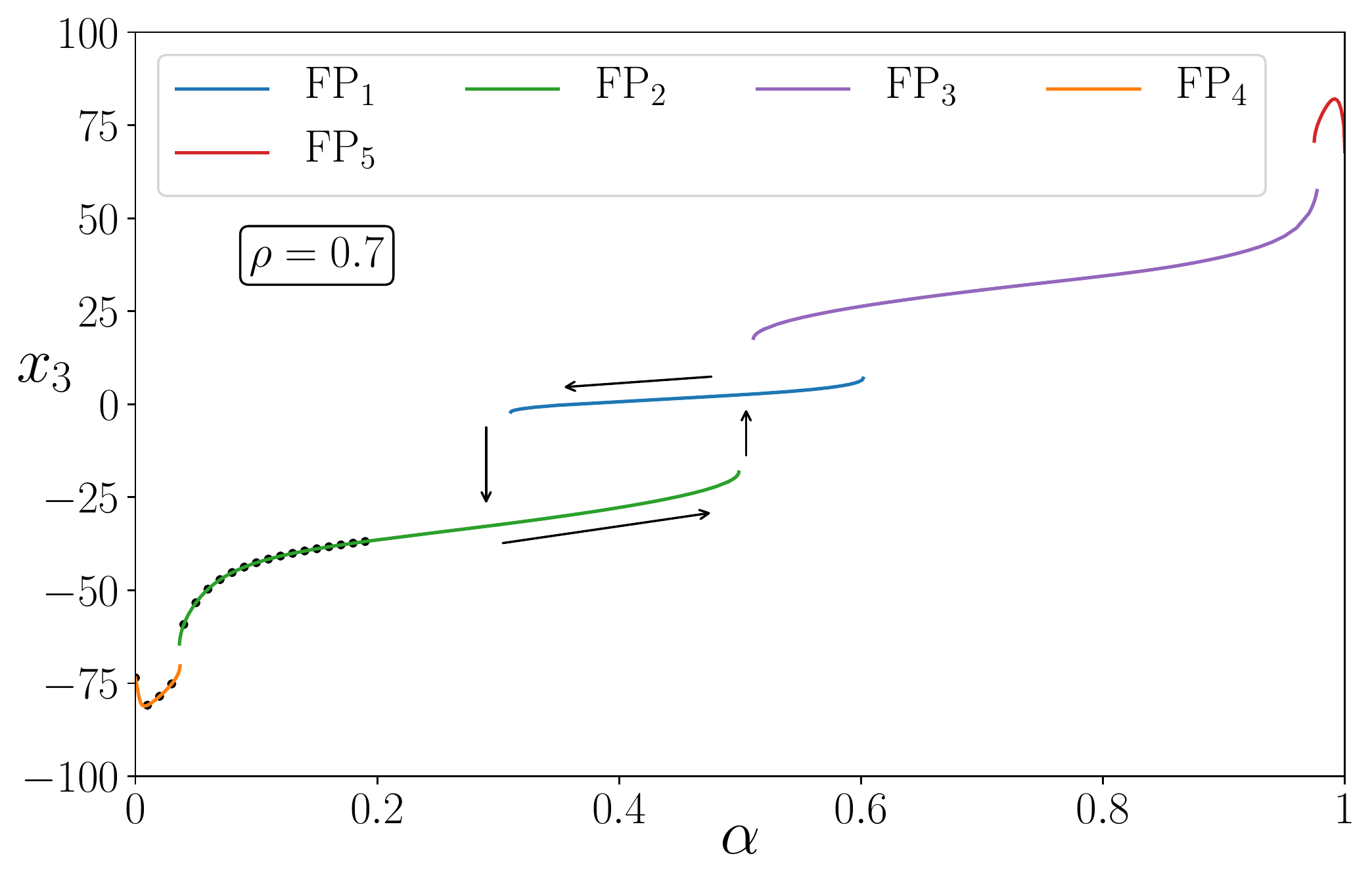}
        \caption{$\rho=0.7$}
        \label{fig:FP_evolution_rho_7}
    \end{subfigure}
    \hfill
    \begin{subfigure}{0.48\textwidth}
        \centering
        \includegraphics[width=\textwidth]{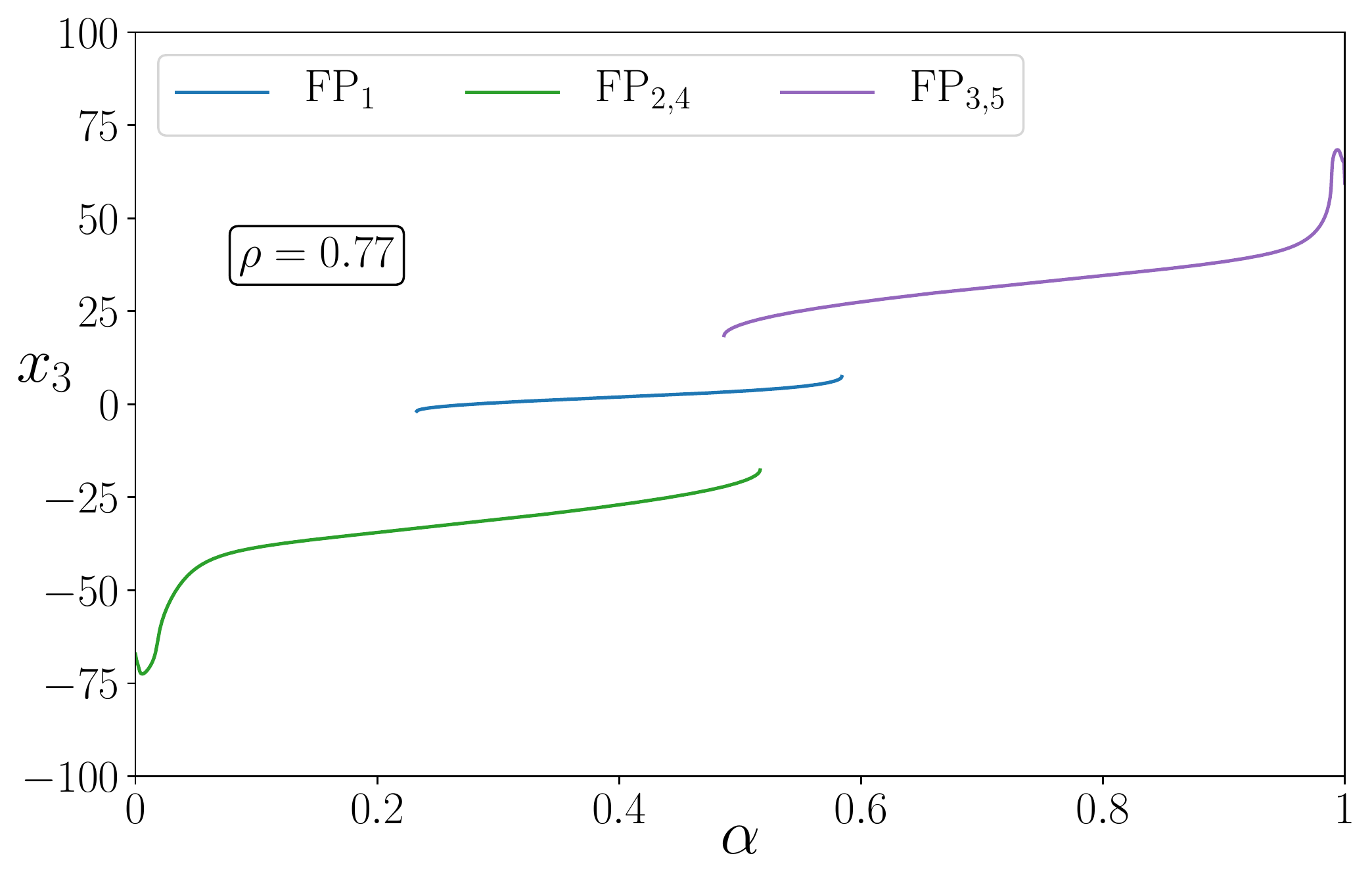}
        \caption{$\rho=0.77$}
        \label{fig:FP_evolution_rho_77}
    \end{subfigure}
    \caption{Evolution of untrained attractors (stable branches of fixed points) in $\left( \alpha, \, x_{3} \right)$-plane for; (a) $\rho = 0.7$ and (b) $\rho = 0.77$.}
    \label{fig:UtrAttTrack}
\end{figure*}

We do this by initialising the state of the RC, trained for a certain $\alpha$ and $\rho$, with one of the corresponding fixed points shown in Fig.\,\ref{fig:rand_FP_ev_rho_7_alpha}. We track the evolution of this fixed point with respect to $\alpha$ by repeating the process of incrementally changing $\alpha$, retraining and initialising the state of the RC with the fixed point corresponding to the previous $\alpha$. The result of this for $\rho = 0.7$ is shown in Fig.\,\ref{fig:FP_evolution_rho_7}.

Here we see the extent of the bistabilities found in Fig.\,\ref{fig:rand_FP_ev_rho_7_alpha}. We also find branches of fixed points and bistabilities closer to the end points of $\alpha$. The evolution with respect to $\alpha$ of the previously found fixed point located in the middle of the chaotic attractors is now labelled as the branch FP$_{1}$ and the evolution of the fixed points closer to $\pzc_{2}$ and $\pzc_{1}$ are labelled as FP$_{2}$ and FP$_{3}$ respectively. We label the evolution of the fixed points closest to $\alpha=0$ as the branch FP$_{4}$ and those closest to $\alpha=1$ as FP$_{5}$. 

Giving rise to these bistabilities are the hysteresis cycles created here. As indicated by the arrows in Fig.\,\ref{fig:FP_evolution_rho_7}, when moving along the FP$_{2}$ branch, as $\alpha$ is increased there is a point where the state of the RC jumps to the FP$_{1}$ branch. After this transition, if $\alpha$ were instead decreased we then remain and continue to track along the FP$_{1}$ branch to the point where the state of the RC returns to the FP$_{2}$ branch, and so on.

Fig.\,\ref{fig:FP_evolution_rho_7} also demonstrates that when the RC fails to reconstruct $\pzc_{1}$ then the predicted trajectory falls onto the FP$_{4}$ and FP$_{2}$ branches. The black points plotted in Fig.\,\ref{fig:FP_evolution_rho_7} are the fixed points that the predictions of $\pzc_{1}$ decay towards in Fig.\,\ref{fig:CaseI_rho_0_7_alpha_left}. As these black points line up directly with the branches of FP$_{4}$ and FP$_{2}$ we conclude that as attractor reconstruction of $\pzc_{1}$ is achieved that these fixed points do not suddenly disappear but that their existence is intrinsic to the dynamics of this RC setup. In modes of failure these branches of fixed points provide routes to stability should the predicted trajectory fall off $\pzc_{1}$. This also gives greater insight to the behaviour of the RC at the boundaries of multifunctionality. It is also important to highlight that the fixed points located at $\alpha=0$ and $1$ also occur in the task specific systems. More specifically, if the RC was successfully trained on data from only $\pzc_{1}$ then within this prediction state space exists the reconstructed attractor $\pzc_{1}$ and the same fixed point, FP$_{5}$, that we find in the multifunctional setup. The same can be said in relation to $\pzc_{2}$ and FP$_{4}$. Furthermore, these untrained attractors have dynamics of their own and interact amongst themselves giving rise to these `behind-the-scenes' bifurcations. 

These stable branches of equilibria would ordinarily be connected by branches of unstable equilibria but given the nature of our approach these cannot be explicitly determined. However, as the end points of these branches are seemingly being drawn together, i.e the right and left end points of both FP$_{4}$ and FP$_{2}$ and likewise for FP$_{3}$ and FP$_{5}$, we infer that this is a signature of the existence of unstable branches of equilibria and evidence that at these end points are saddle-node (SN) bifurcations. To help strengthen this claim, we increase $\rho$ to $0.77$ and track the evolution of the fixed points as before. The result of this is shown in Fig.\,\ref{fig:FP_evolution_rho_77} where the right and left end points of FP$_{4}$ and FP$_{2}$ as well as FP$_{3}$ and FP$_{5}$ have connected together resulting in two branches of fixed points which we call, FP$_{2,4}$ and FP$_{3,5}$. This particular behaviour is indicative of two cusp bifurcations taking place about $\alpha \approx 0.034$ and $0.988$ as $\rho$ is increased from $0.7$ to $0.77$. In addition, we see a region of `tristability' here for $0.4867 \leq \alpha \leq 0.5165$ where there is an overlap between all three branches.

\begin{figure*}
    \centering
    \includegraphics[width=0.99\textwidth]{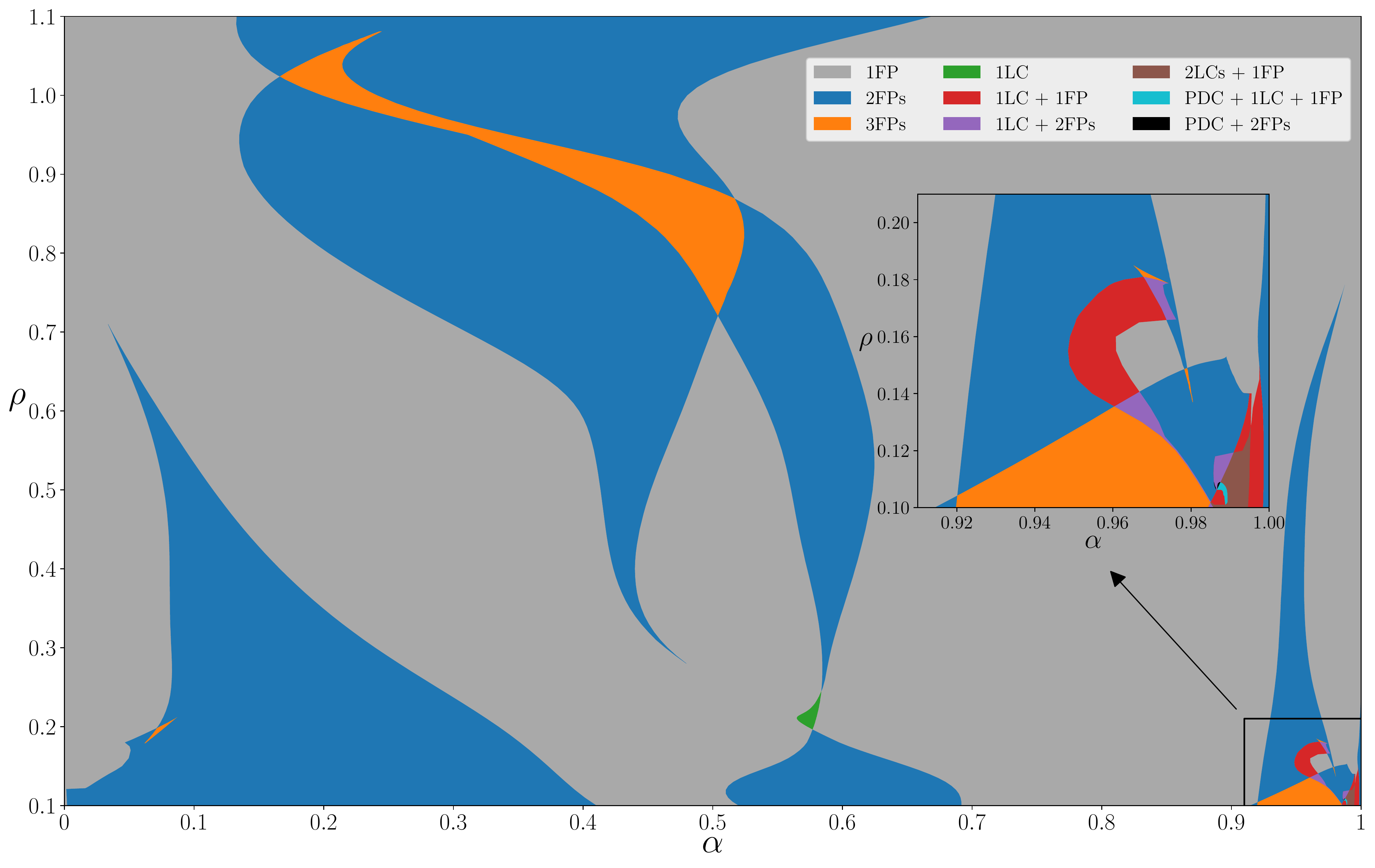}
    \caption{Classification of `Untrained Attractors' in the $\left( \alpha, \, \rho \right)$-plane. \\(FP: Fixed Point, LC: Limit Cycle, PDC: Period Doubling Cascade)}
    \label{fig:UntrainedAtts}
\end{figure*}

We continue exploring and characterising the behaviour of these untrained attractors by tracking their evolution in the $\left( \alpha, \, \rho \right)$-plane. The result of this is depicted in Fig.\,\ref{fig:UntrainedAtts}.

In this picture we see the previously mentioned cusp bifurcations taking place at $\left( \alpha, \, \rho \right) = \left( 0.0337, \, 0.71 \right)$ and $\left( 0.9875, \, 0.76 \right)$. We also find a cusp bifurcation of the FP$_{1}$ and FP$_{2}$ branches taking place at $\left( \alpha, \, \rho \right) \approx \left( 0.48, \, 0.27 \right)$. We denote FP$_{1,2}$ as the branch of stable fixed points emerging from this cusp bifurcation.

\begin{figure}
    \centering
    \includegraphics[width=0.48\textwidth]{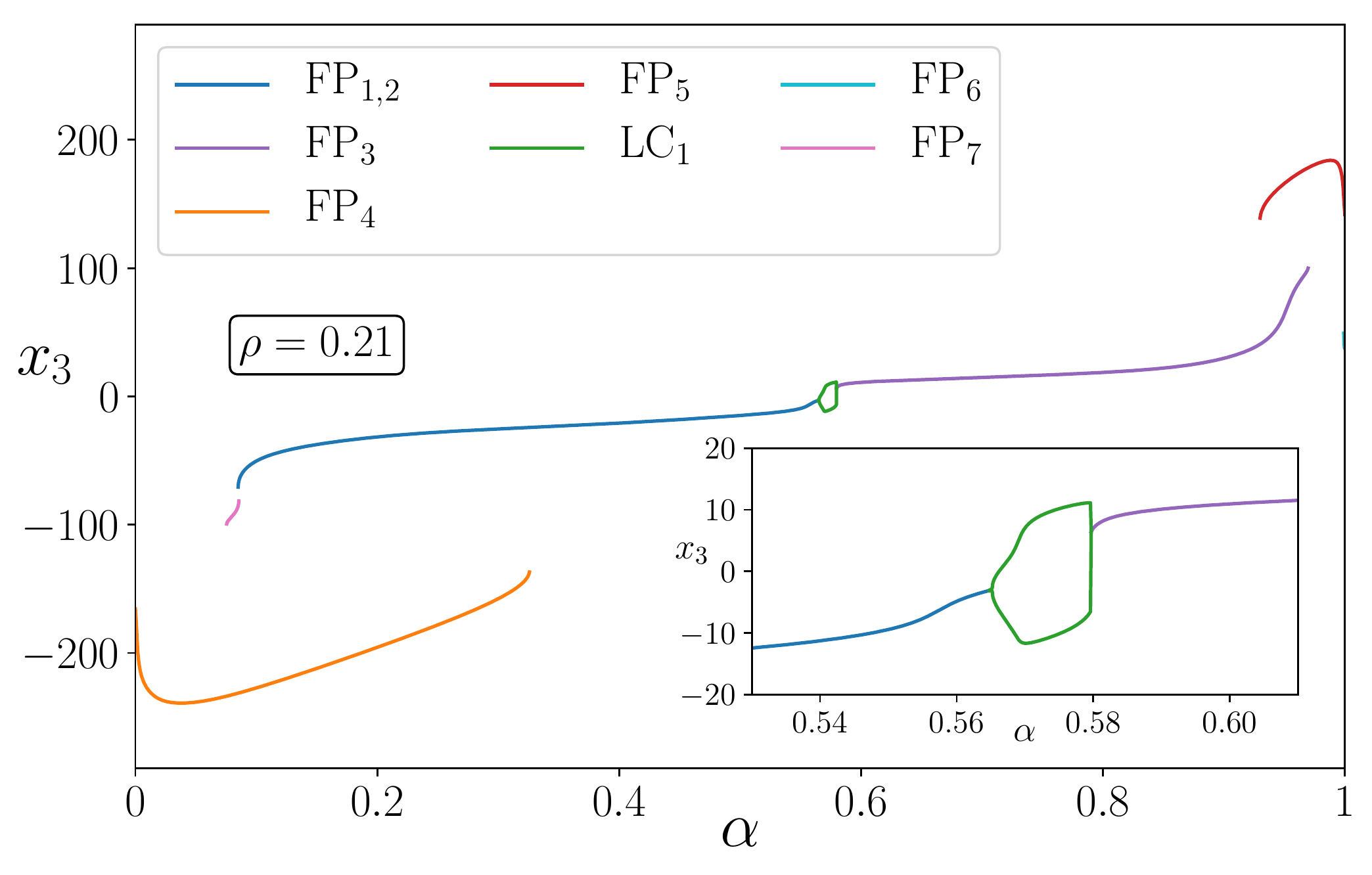}
    \caption{Tracking Untrained Attractors for $\rho = 0.21$: Max and min values of limit cycle, LC$_{1}$, born at the right and left end points of the FP$_{1,2}$ and FP$_{3}$ branches. Also seen are the FP$_{4}$, FP$_{5}$, FP$_{6}$ and FP$_{7}$, branches of stable fixed points.}
    \label{fig:FP_evol_rho_021}
\end{figure}

Fig.\,\ref{fig:UntrainedAtts} also reveals the existence of several limit cycles. As $\rho$ is decreased from $0.27$, the bistability between FP$_{3}$ and FP$_{1,2}$ is lost at $\left( \alpha, \, \rho \right) \approx \left( 0.5837, \, 0.2449 \right)$. As these branches begin to drift apart, we show in Fig.\,\ref{fig:FP_evol_rho_021} for $\rho = 0.21$ that a period-1 limit cycle, LC$_{1}$, is born in this gap. Plotted here is the evolution of the maximum and minimum values of LC$_{1}$ versus $\alpha$. As $\rho$ is decreased further, the bistability between the FP$_{3}$ and FP$_{1,2}$ branches resumes from $\left( \alpha, \, \rho \right) \approx \left( 0.577, \, 0.1965 \right)$ resulting in the death of LC$_{1}$.

We also see a relatively small region of tristability between FP$_{1,2}$, FP$_{7}$ and FP$_{4}$ in Fig.\,\ref{fig:FP_evol_rho_021}. While the left end of FP$_{1,2}$ had at one time appeared to be growing towards and potentially connecting to FP$_{4}$ in another cusp bifurcation, instead it has broken off due to a cusp bifurcation taking place on FP$_{1,2}$ at $\left( \alpha, \, \rho \right) \approx \left( 0.212, \, 0.087 \right)$ resulting in a new branch of fixed points FP$_{7}$ as seen in Fig.\,\ref{fig:FP_evol_rho_021}. The existence of FP$_{7}$ is relatively brief as its own end points are quickly drawn together and disappear entirely at $\left( \alpha, \, \rho \right) \approx \left( 0.062, \, 0.179 \right)$ in Fig.\,\ref{fig:UntrainedAtts}. This particular behaviour also occurs on FP$_{3}$ where we will later provide a more detailed picture and explanation of these events. We also find the beginnings of a branch of stable fixed points emerging from the right hand side of Fig.\,\ref{fig:FP_evol_rho_021} which we label as FP$_{6}$. Later we will discuss some interesting properties emanating from this branch.

\begin{figure}
    \centering
    \includegraphics[width=0.48\textwidth]{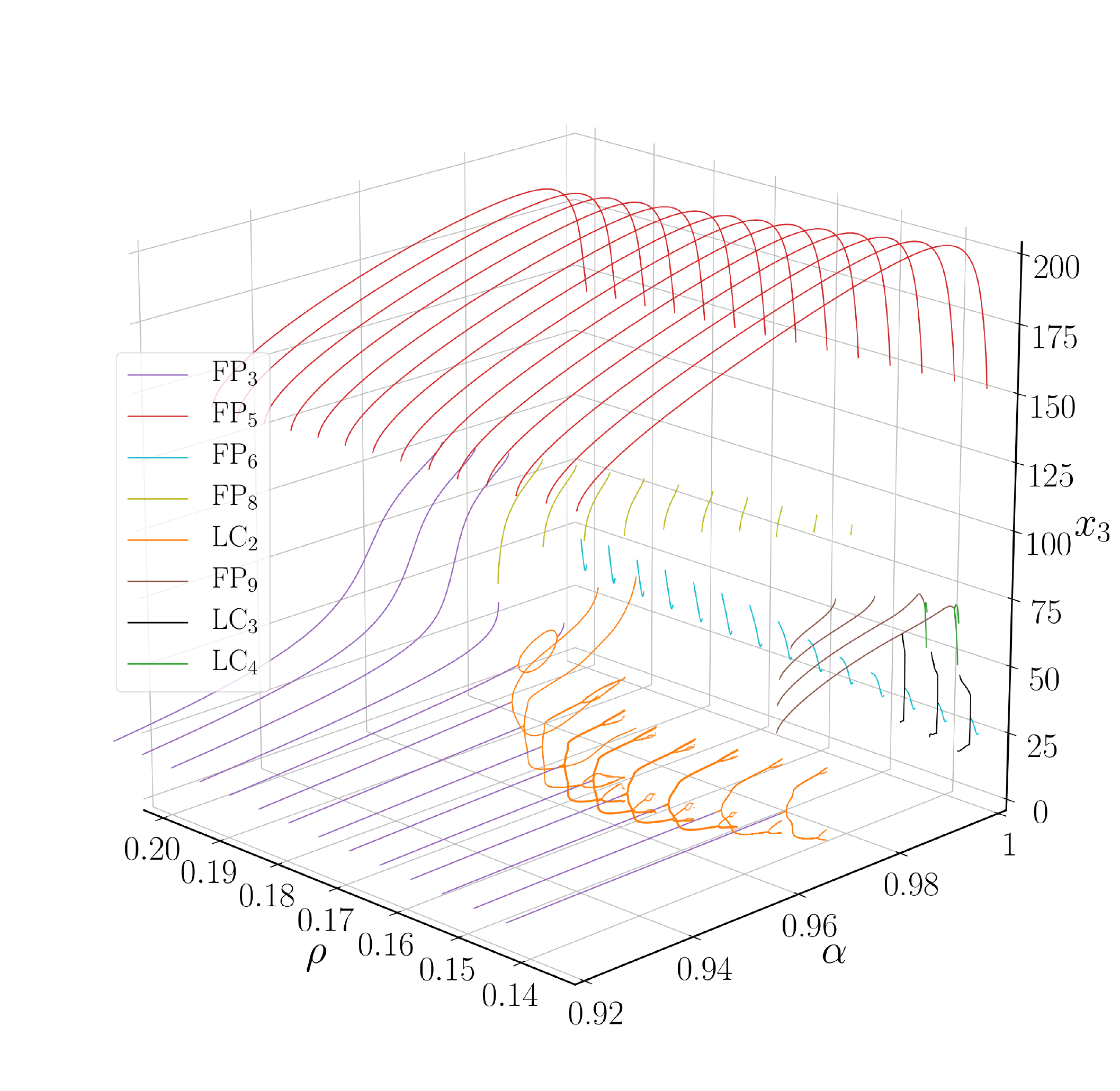}
    \caption{Behaviour of the Untrained Attractors in the predicted $x_{3}$ direction for large $\alpha$ and small $\rho$}
    \label{fig:swallow}
\end{figure}

Given the abundant variety of behaviour exhibited by the untrained attractors for large $\alpha$ and small $\rho$, as seen in the inset plot of Fig.\,\ref{fig:UntrainedAtts}, we provide a more expansive picture in Fig.\,\ref{fig:swallow} depicting how certain bifurcations arise.

The cusp bifurcation taking place at $\left( \alpha, \rho \right) = \left( 0.965, 0.185 \right)$ gives rise to a relatively small tristable region. This cusp bifurcation originates from a buckling of the FP$_{3}$ branch where a new branch of stable fixed points emerges, labelled as FP$_{8}$ in Fig.\,\ref{fig:swallow}. 

A limit cycle is born at the right end point of FP$_{3}$ for $\rho \approx 0.181$. In Fig.\,\ref{fig:swallow} we plot the evolution of the maximum and minimum $x_{3}$-values of this period-1 limit cycle which we label as LC$_{2}$. Close to this cusp bifurcation we see for $\rho = 0.18$, that LC$_{2}$ is contained within a Hopf-bubble as it exists between two supercritical Hopf bifurcations. However, as $\rho$ is decreased further, the amplitude of oscillation of LC$_{2}$ increases and the bubble bursts. We also find that LC$_{2}$ undergoes period-doubling (PD) bifurcations, the first of which was found nearby $\left( \alpha, \rho \right) = \left( 0.959, 0.166 \right)$. These PD bifurcations attribute to the significant reduction in $\alpha$-values for which we are able to track LC$_{2}$. Also shown in Fig.\,\ref{fig:swallow} are cases where LC$_{2}$ subsequently encounters another PD bifurcation giving rise to a period-4 limit cycle.

The existence of FP$_{8}$ is relatively short, as $\rho$ is decreased its end points are drawn closer together until this branch becomes but a single point-like attractor at $\left( \alpha, \rho \right) \approx \left( 0.98, 0.137 \right)$ and we are no longer able to track. This event and that mentioned earlier regarding FP$_{7}$ is evidence that further flavours of cusp-like bifurcations take place in the prediction state space. A codimension-3 bifurcation analysis (involving either $\sigma$, $\beta$ or $\gamma$) could, for example, unveil a swallowtail bifurcation point (the point at which two cusp branches collide). For reading on more cusp-like bifurcations see \textcite{kuznetsov2013elements} or \textcite{guckenheimer13_book}. Alternatively, if swallowtail bifurcation points were found to take place in lower dimensional representations of Eq.\,\eqref{PredRes} (for $N=2,3,\ldots$), then their existence could be generalised to the higher dimensional picture. The benefit of reducing the dimension would allow for the use of numerical continuation software like AUTO \cite{AUTO_Doedel} on Eq.\,\eqref{PredRes}. Moreover, this approach facilitates the study of unstable attractors and in identifying further dynamical features inherent to Eq.\,\eqref{PredRes}. We leave this for future work.

As shown in Fig.\,\ref{fig:swallow}, we find another branch of fixed points which we label as FP$_{9}$. For $\rho = 0.135$ we see here that at the right end point of FP$_{9}$ a limit cycle labelled as LC$_{4}$ is born. Increasing $\rho$ results in the loss of LC$_{4}$ as the end points of FP$_{9}$ are being drawn closer together where eventually at $\left( \alpha, \rho \right) = \left( 0.989, 0.153 \right)$ we are no longer able to track.

\begin{figure}
    \centering
    \hfill
    \begin{subfigure}{0.48\textwidth}
        \centering
        \includegraphics[width=0.99\textwidth]{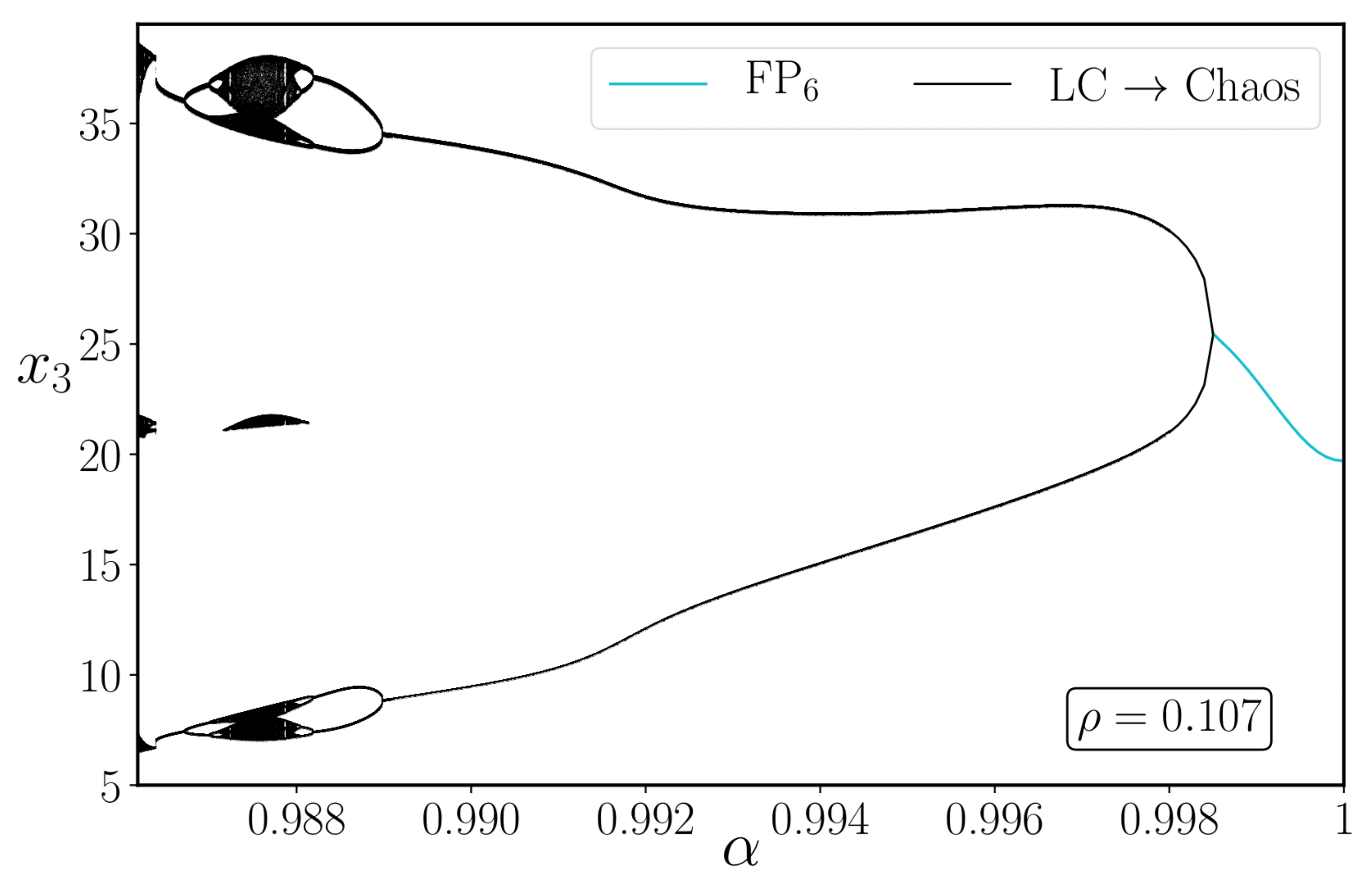}
        \label{fig:Bif_PDoubling}
    \end{subfigure}
    \hfill
    \vskip\baselineskip
    \begin{subfigure}{0.48\textwidth}
        \centering
        \includegraphics[width=0.99\textwidth]{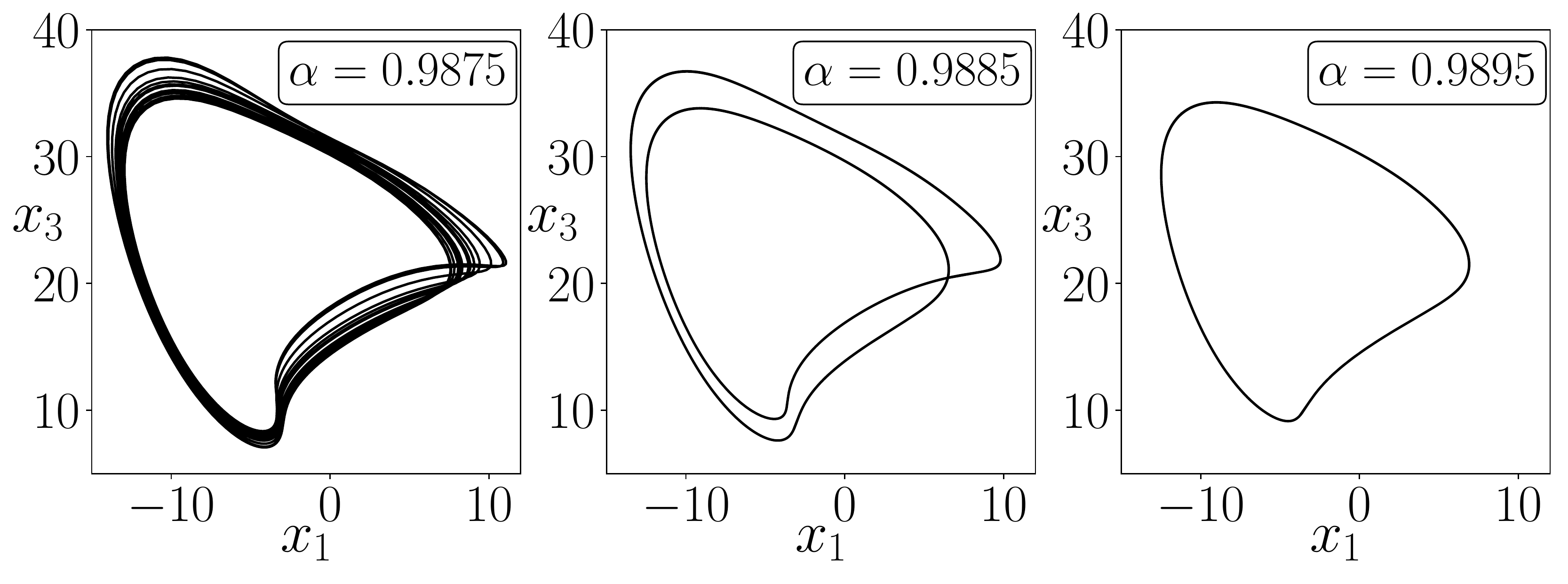}
        \label{fig:LCtoC}
    \end{subfigure}
    \caption{($\rho = 0.107$): Evolution of FP$_{6}$ as it transition to a limit cycle and then to a chaotic attractor through an infinite sequence of period-doubling bifurcations. Also shown are snapshots of this limit cycle along its route to chaos.}
    \label{fig:PDoublingToChaos}
\end{figure}

Throughout Fig.\,\ref{fig:swallow} we plot the evolution of the previously mentioned branch, FP$_{6}$. As $\rho$ is decreased, a period-1 limit cycle, labelled as LC$_{3}$, is born from the left end point of FP$_{6}$. The brown region in Fig.\,\ref{fig:UntrainedAtts} depicts the coexistence of LC$_{3}$, LC$_{4}$, and FP$_{5}$. However, as we follow the evolution of the local maxima and minima of LC$_{3}$ there are certain points in the $\left( \alpha, \, \rho \right)$-plane where it undergoes a PD bifurcation. Furthermore, there at times at which one PD bifurcation leads to another and in turn triggers a period-doubling cascade (PDC) ultimately resulting in chaotic behaviour. We find two relatively small distinct regions in Fig.\,\ref{fig:UntrainedAtts} (coloured in black and cyan) where PD bifurcations of LC$_{3}$ lead to PDCs. An example of this particular sequence of PD bifurcations is shown in Fig.\,\ref{fig:PDoublingToChaos} when setting $\rho = 0.107$.

Starting from $\alpha = 1$, we see in Fig.\,\ref{fig:PDoublingToChaos} how FP$_{6}$ evolves as $\alpha$ is decreased. We plot the local maxima and minima of LC$_{3}$ sprouting from the left end point of FP$_{6}$ for $\alpha = 0.9986$ and continue to track as $\alpha$ is reduced further. Here we see the first of these PD bifurcations occurring at $\alpha \approx 0.989$ where there are now two distinct local maxima and minima of LC$_{3}$. This particular behaviour is depicted below the bifurcation diagram in Fig.\,\ref{fig:PDoublingToChaos} with snapshots of LC$_{3}$ as it travels along its route to a PDC in the $\left( x_{1}, \, x_{3} \right)$ prediction state space. Here, we see how LC$_{3}$ transitions from period-1 at $\alpha = 0.9895$ to period-2 at $\alpha = 0.9885$. Decreasing $\alpha$ further to $0.9875$ we see that there has been an infinite number of PD bifurcations giving rise to a chaotic attractor. Periodic behaviour briefly resumes for decreasing $\alpha$ further before entering another PDC as a second bout of chaos begins at $\alpha = 0.98618$ and ends at $0.98638$.

Despite that multifunctionality is not achieved in the relatively small regions in which we find these PDCs it is a remarkable result as it shows that when attempting to train the RC to promote a coexistence of two desired chaotic attractors it is also possible for another chaotic attractor to exist in the background. Moreover, for a different choice of the other reservoir parameters, these PDCs could play a larger role in the prediction state space and further influence the RCs ability to reconstruct a given attractor. Throughout our results we have kept the topology of the RC matrices, $\textbf{M}$, and $\textbf{W}_{in}$ fixed. Given a different initialisation of these matrices we expect that quantitative changes in these bifurcation figures would occur but that the main characteristics would remain.

Overall, these results indicate that regardless of the particular attractor one is attempting to reconstruct, there will always be some untrained attractor present in the prediction state space. The long term characterisation of the RCs prediction of a given attractor in Figs.\,\ref{fig:beta_2_rho_alpha_err_att1}-\ref{fig:beta_2_rho_alpha_err_att2} combined with the classification of the untrained attractors in Fig.\,\ref{fig:UntrainedAtts} provides us with a clearer picture of the prediction state space for a given $\alpha$ and $\rho$. When put together we are able to see the regions in which multifunctionality was achieved, the manner in which the prediction fails, and the particular untrained attractor it can tend towards. Moreover, one could argue that attractor reconstruction may fail if the basin of attraction of the desired attractor interferes with one of these untrained attractors.

\section{\label{sec:DisConc}Conclusion}

In this paper we have demonstrated that a Reservoir Computer (RC) can be trained to exhibit multifunctionality whereby, for a given initial condition, the climate of more than one chaotic attractor can be successfully reconstructed in its prediction state space. 

In order to train a RC to express multifunctionality we introduce the `blending technique' as a means to combine and weight data from two different sources. We test the flexibility of this technique by training the RC to reconstruct a coexistence of chaotic attractors from; a system which already exhibits multistability, two parameter settings of a system and, two different systems entirely. 

However, in order to achieve the desired outcome, there is a crucial dependence on certain reservoir and training parameters, in particular the `blending' parameter, $\alpha$, and the spectral radius of the reservoirs internal connection matrix, $\rho$. When varying these parameters, we find an abundance of nontrivial transitions between multifunctionality and modes of failure as there is a competition between attractors in the prediction state space of the RC for a given $\alpha$ and $\rho$. For example, when attempting to reconstruct the climate of a given chaotic attractor, there are times where the prediction can switch to the other. In this case, the RC is able to reconstruct the climate of only one chaotic attractor. This behaviour comes as a consequence of training a RC to reconstruct the climate of more than one attractor. Furthermore, we see that in the event of failure, the predicted trajectory on a given chaotic attractor can tend toward a limit cycle or fixed point. 

On closer inspection, we find that even when the RC is successfully trained to express multifunctionality there are additional attractors existing within the prediction state space that were not involved in the training, we call these the `untrained attractors'. Moreover, it is shown that these untrained attractors have dynamics of their own and play a significant role in the event that the desired attractor cannot be successfully reconstructed.

By tracking the evolution of these untrained attractors with respect to $\alpha$ and $\rho$ we have identified a number of `behind-the-scenes' bifurcations. In particular when tracking the evolution of FP$_{6}$ we see in Fig.\,\ref{fig:PDoublingToChaos} how this branch of stable fixed points transitions to the limit cycle LC$_{3}$ which subsequently undergoes a series of period-doubling bifurcations to the point of provoking a period-doubling cascade inevitably leading towards the creation of a chaotic attractor. So while we try to train the RC to reconstruct a coexistence of two specific chaotic attractors in its prediction state space there is another chaotic attractor created in a sub rosa fashion.

Much like our investigation of the untrained attractors in the RCs prediction state space, there is also evidence of similarly occurring events in the brain which influence and disrupt its normal behaviour. It is understood that neurological disorders like Parkinson's disease or Epilepsy occur as a result of certain active regions of the brain deviating from its normal state of operation and then becoming trapped within some undesirable behaviour beyond which it may not be able to resume its normal function. The effort is often made to model and control these events in order to counteract the effects of these illnesses \cite{TassKurths98,RosenblumPikovsky04,Tass06NLDyn}.

The ability to combine attractors from different sources to then coexist in the same state space broadens the current set of RC applications. This demonstrates that instead of having to change parameters in a system for it to exhibit a different behaviour, the various desired modes of operation can coexist in the state space of a multifunctional RC where an appropriate controller could in theory be designed to switch between attractors \cite{richter02controlchaos}. This draws further parallels to biological neural networks, where such a control mechanism is comparable to neuromodulators \cite{Harris91modulation}, a hierarchical system of neurons which some believe to be involved in instigating the switching of activity patterns in multifunctional neural networks. Additionally, we have illustrated that these input sources are not limited to the one system. We show this in Fig.\,\ref{fig:3D_CaseIII} where the RC was successfully trained to permit the coexistence of the Lorenz butterfly attractor, $\pzl$, and the chaotic attractor $\pzc_{2}$ generated from Eq.\,\eqref{3DSys}. Naturally the question emerges as to the amount of attractors that can be successfully trained to coexist in the RCs prediction state space, we leave this for future work.

The results of this paper emanate from employing the dyadic `two-way street' approach of framing neurological features in the context of dynamical systems. Moreover, this work indicates that if other hypotheses or known facets of the brain can be articulated in this manner then there lies the potential to portray it artificially.

\begin{acknowledgments}
This work was funded by the Irish Research Council Enterprise Partnership Scheme (Grant No. EPSPG/2017/301). A.\,F.\, would like to thank Sebastian Wieczorek for introducing him to Reservoir Computing, Paul O' Keeffe and Christopher O' Connor for their helpful conversations.
\end{acknowledgments}

\section*{DATA AVAILABILITY}
The data that support the findings of this study are available from the corresponding author upon reasonable request.




\end{document}